\pdfoutput=1
\documentclass{article}
 
\makeatletter
 
\usepackage[verbose=true,letterpaper]{geometry}
\AtBeginDocument{
  \newgeometry{
    textheight=9in,
    textwidth=6.5in,
    top=1in,
    headheight=14pt,
    headsep=25pt,
    footskip=30pt
  }
}
 
\newcommand{\headeright}{A Preprint}
\newcommand{\undertitle}{A Preprint}
\newcommand{\shorttitle}{\@title}
\renewcommand{\shorttitle}{Edge-Deployable LDW \& TSD/TSR Framework}
\renewcommand{\undertitle}{%
Published in \emph{Machine Vision and Applications}, Vol.~37, Article 61 (2026)\\
Publisher: Springer Nature \quad $\vert$ \quad Published: 17 April 2026\\
DOI: \href{https://doi.org/10.1007/s00138-026-01816-x}{10.1007/s00138-026-01816-x}
}
\usepackage{fancyhdr}
\fancyheadoffset{0pt}
\def\keywordname{{\bfseries \emph{Keywords}}}%
\def\keywords#1{\par\addvspace\medskipamount{\rightskip=0pt plus1cm
\def\and{\ifhmode\unskip\nobreak\fi\ $\cdot$
}\noindent\keywordname\enspace\ignorespaces#1\par}}
 
\renewcommand{\normalsize}{%
  \@setfontsize\normalsize\@xpt\@xipt
  \abovedisplayskip      7\p@ \@plus 2\p@ \@minus 5\p@
  \abovedisplayshortskip \z@ \@plus 3\p@
  \belowdisplayskip      \abovedisplayskip
  \belowdisplayshortskip 4\p@ \@plus 3\p@ \@minus 3\p@
}
\normalsize
\renewcommand{\small}{%
  \@setfontsize\small\@ixpt\@xpt
  \abovedisplayskip      6\p@ \@plus 1.5\p@ \@minus 4\p@
  \abovedisplayshortskip \z@  \@plus 2\p@
  \belowdisplayskip      \abovedisplayskip
  \belowdisplayshortskip 3\p@ \@plus 2\p@   \@minus 2\p@
}
\renewcommand{\footnotesize}{\@setfontsize\footnotesize\@ixpt\@xpt}
\renewcommand{\scriptsize}{\@setfontsize\scriptsize\@viipt\@viiipt}
\renewcommand{\tiny}{\@setfontsize\tiny\@vipt\@viipt}
\renewcommand{\large}{\@setfontsize\large\@xiipt{14}}
\renewcommand{\Large}{\@setfontsize\Large\@xivpt{16}}
\renewcommand{\LARGE}{\@setfontsize\LARGE\@xviipt{20}}
\renewcommand{\huge}{\@setfontsize\huge\@xxpt{23}}
\renewcommand{\Huge}{\@setfontsize\Huge\@xxvpt{28}}
 
\providecommand{\section}{}
\renewcommand{\section}{%
  \@startsection{section}{1}{\z@}%
                {-2.0ex \@plus -0.5ex \@minus -0.2ex}%
                { 1.5ex \@plus  0.3ex \@minus  0.2ex}%
                {\large\bf\raggedright}%
}
\providecommand{\subsection}{}
\renewcommand{\subsection}{%
  \@startsection{subsection}{2}{\z@}%
                {-1.8ex \@plus -0.5ex \@minus -0.2ex}%
                { 0.8ex \@plus  0.2ex}%
                {\normalsize\bf\raggedright}%
}
\providecommand{\subsubsection}{}
\renewcommand{\subsubsection}{%
  \@startsection{subsubsection}{3}{\z@}%
                {-1.5ex \@plus -0.5ex \@minus -0.2ex}%
                { 0.5ex \@plus  0.2ex}%
                {\normalsize\bf\raggedright}%
}
\providecommand{\paragraph}{}
\renewcommand{\paragraph}{%
  \@startsection{paragraph}{4}{\z@}%
                {1.5ex \@plus 0.5ex \@minus 0.2ex}%
                {-1em}%
                {\normalsize\bf}%
}
\providecommand{\subparagraph}{}
\renewcommand{\subparagraph}{%
  \@startsection{subparagraph}{5}{\z@}%
                {1.5ex \@plus 0.5ex \@minus 0.2ex}%
                {-1em}%
                {\normalsize\bf}%
}

\newlength{\@abovecaptionskip}
\newlength{\@belowcaptionskip}
 
\renewenvironment{table}
  {\setlength{\abovecaptionskip}{\@belowcaptionskip}%
   \setlength{\belowcaptionskip}{\@abovecaptionskip}%
   \@float{table}}
  {\end@float}
 
\renewcommand{\footnoterule}{\kern-3\p@ \hrule width 12pc \kern 2.6\p@}
\def\@listi  {\leftmargin\leftmargini}
\def\@listii {\leftmargin\leftmarginii
              \labelwidth\leftmarginii
              \advance\labelwidth-\labelsep
              \topsep  2\p@ \@plus 1\p@    \@minus 0.5\p@
              \parsep  1\p@ \@plus 0.5\p@ \@minus 0.5\p@
              \itemsep \parsep}
\def\@listiii{\leftmargin\leftmarginiii
              \labelwidth\leftmarginiii
              \advance\labelwidth-\labelsep
              \topsep    1\p@ \@plus 0.5\p@ \@minus 0.5\p@
              \parsep    \z@
              \partopsep 0.5\p@ \@plus 0\p@ \@minus 0.5\p@
              \itemsep \topsep}
\def\@listiv {\leftmargin\leftmarginiv
              \labelwidth\leftmarginiv
              \advance\labelwidth-\labelsep}
\def\@listv  {\leftmargin\leftmarginv
              \labelwidth\leftmarginv
              \advance\labelwidth-\labelsep}
\def\@listvi {\leftmargin\leftmarginvi
              \labelwidth\leftmarginvi
              \advance\labelwidth-\labelsep}
 
\providecommand{\maketitle}{}
\renewcommand{\maketitle}{%
  \par
  \begingroup
    \renewcommand{\thefootnote}{\fnsymbol{footnote}}
    \long\def\@makefntext##1{%
      \parindent 1em\noindent
      \hbox to 1.8em{\hss $\m@th ^{\@thefnmark}$}##1
    }
    \thispagestyle{empty}
    \@maketitle
    \@thanks
  \endgroup
  \let\maketitle\relax
  \let\thanks\relax
}
 
\newcommand{\@toptitlebar}{
  \hrule height 2\p@
  \vskip 0.25in
  \vskip -\parskip%
}
\newcommand{\@bottomtitlebar}{
  \vskip 0.29in
  \vskip -\parskip
  \hrule height 2\p@
  \vskip 0.09in%
}
 
\providecommand{\@maketitle}{}
\renewcommand{\@maketitle}{%
  \vbox{%
    \hsize\textwidth
    \linewidth\hsize
    \vskip 0.1in
    \@toptitlebar
    \centering
    {\LARGE\sc \@title\par}
    \@bottomtitlebar
    \textsc{\undertitle}\\
    \vskip 0.1in
    \def\And{%
      \end{tabular}\hfil\linebreak[0]\hfil%
      \begin{tabular}[t]{c}\bf\rule{\z@}{24\p@}\ignorespaces%
    }
    \def\AND{%
      \end{tabular}\hfil\linebreak[4]\hfil%
      \begin{tabular}[t]{c}\bf\rule{\z@}{24\p@}\ignorespaces%
    }
    \begin{tabular}[t]{c}\bf\rule{\z@}{24\p@}\@author\end{tabular}%
  \vskip 0.4in \@minus 0.1in \center{\@date}   \vskip 0.2in
  }
}
 
\newcommand{\ftype@noticebox}{8}
\newcommand{\@notice}{%
  \enlargethispage{2\baselineskip}%
  \@float{noticebox}[b]%
    \footnotesize\@noticestring%
  \end@float%
}
 
\renewenvironment{abstract}
{
  \centerline
  {\large \bfseries \scshape Abstract}
  \begin{quote}
}
{
  \end{quote}
}
 
\makeatother
 
\usepackage[utf8]{inputenc}
\usepackage[T1]{fontenc}
\usepackage{hyperref}
\usepackage{url}
\usepackage{microtype}
\usepackage{graphicx}
\usepackage[numbers]{natbib}
\usepackage{doi}
 
\usepackage{amsmath,amsfonts,amssymb}
\usepackage{nicefrac}
 
\usepackage{newunicodechar}
\newunicodechar{≈}{\ensuremath{\approx}}
\usepackage{bbding}          
 
\usepackage[ruled,vlined]{algorithm2e}
 
\usepackage{booktabs}
\usepackage{dcolumn}
\usepackage{multirow}
\usepackage{array}
 
\usepackage{caption}
\usepackage[caption=false,font=normalsize,labelfont=sf,textfont=sf]{subfig}
\usepackage{textcomp}
 
\usepackage{enumitem}
\usepackage{xcolor}
\usepackage{soul}
 
\title{Edge Deployable Lightweight Framework for Real-Time Driver Safety System with Integrated Lane Departure and Traffic Sign Recognition}

\author{
  Hassaan Ali Qureshi \\
  Department of Mechatronics Engineering \\
  College of Electrical and Mechanical Engineering \\
  National University of Sciences and Technology \\
  Islamabad 44000, Pakistan \\
  \And
  Danish Hussain\thanks{Corresponding author: \texttt{danish.hussain@smme.nust.edu.pk}} \\
  Department of Robotics and Artificial Intelligence \\
  School of Mechanical and Manufacturing Engineering \\
  National University of Sciences and Technology \\
  Islamabad 44000, Pakistan \\
  \And
  Ali Hassan\thanks{Corresponding author: \texttt{alihassan@ceme.nust.edu.pk}} \\
  Department of Computer and Software Engineering \\
  College of Electrical and Mechanical Engineering \\
  National University of Sciences and Technology \\
  Islamabad 44000, Pakistan \\
  \And
  Aon Safdar \\
  School of Computer Science \\
  University College Dublin \\
  Dublin, Ireland \\
  \And
  Anas Bin Aqeel \\
  Department of Mechatronics Engineering \\
  College of Electrical and Mechanical Engineering \\
  National University of Sciences and Technology \\
  Islamabad 44000, Pakistan \\
  \And
  Muhammad Rafique \\
  School of Energy Science and Engineering \\
  Chongqing Research Institute of Harbin Institute of Technology \\
  Chongqing 150001, PR China \\
  \And
  Zohaib Riaz \\
  Department of Mechatronics Engineering \\
  College of Electrical and Mechanical Engineering \\
  National University of Sciences and Technology \\
  Islamabad 44000, Pakistan \\
  \And
  Farhan Hussain \\
  Department of Computer and Software Engineering \\
  College of Electrical and Mechanical Engineering \\
  National University of Sciences and Technology \\
  Islamabad 44000, Pakistan \\
}

\renewcommand{\shorttitle}{Edge-Deployable LDW \& TSD/TSR Framework}

\hypersetup{
pdftitle={Edge Deployable Lightweight Framework for Real-Time Driver Safety System with Integrated Lane Departure and Traffic Sign Recognition},
pdfsubject={cs.CV, cs.RO},
pdfauthor={Hassaan Ali Qureshi, Danish Hussain, Ali Hassan, Aon Safdar, Anas Bin Aqeel, Muhammad Rafique, Zohaib Riaz, Farhan Hussain},
pdfkeywords={Advanced Driver Assistance Systems, Lane Departure Warning, Traffic Sign Recognition, Edge Computing, Real-Time Implementation},
}

\begin{document}
\maketitle

\begin{abstract}
Lane departure and traffic-sign negligence are leading contributors to road accidents, particularly in regions with non-standardized signage and limited annotated data. Existing Advanced Driver Assistance Systems (ADAS) solutions are often developed as disjointed submodules, are rarely validated for real-time operation, and struggle to balance accuracy with low-latency inference for embedded deployment. We address these gaps with a unified, edge-deployable ADAS framework that simultaneously performs lane-departure warning (LDW) and traffic-sign detection and recognition (TSD/TSR), and validate the proposed system on automotive-class devices. Our framework combines a lightweight, lane-segmentation-driven LDW, which includes a drivable-area fallback for degraded markings, with a two-stage TSD/TSR pipeline consisting of efficient sign localization followed by fine-grained classification, designed to preserve accuracy at embedded latency. To support region-specific deployment under data scarcity, we curate a 35-class local traffic-sign dataset and introduce a generalized augmentation and balancing profile, expanding the corpus from 359 images to 8,750 samples. On an NVIDIA Jetson Orin Nano, the integrated system sustains 15 FPS end to end (32 FPS for LDW and 18 FPS for TSD/TSR). The TSD/TSR branch achieves 97.4 mAP for detection and 95.6\% validation accuracy (93.1\% on a local real-world set) while maintaining an approximately 13.9 M parameter footprint. Relative to recent methods that target a single task or report results only on desktop-class GPUs, our approach delivers both LDW and TSD/TSR with real-time performance on edge hardware, providing a practical pathway to low-cost, field-ready ADAS in resource-constrained settings.
\end{abstract}


\vspace{0.5em}
\noindent\textbf{Code and data availability:} \textbf{https://anonymous.4open.science/r/LDW-TSDR-062D/}

\section{Introduction}
Enhancing road safety and assisting drivers in potentially hazardous situations are some of the benefits of advanced driver assistance systems. These systems are designed to reduce human error, a primary cause of road accidents, by incorporating intelligent features such as collision avoidance, lane keeping, and traffic sign recognition. The Lane Departure Warning System (LDWS) and Traffic Sign Recognition enable detection and interpretation of road signs, which is a critical component of ADAS \cite{neumann2024analysis}. 

Deep learning has been used to improve lane detection. For instance, Han et al. proposed LLDNet, a lightweight model with just 0.26 million parameters achieve 96.31\% accuracy in lane detection \cite{khan2022lldnet}.  Luo et al. introduced LATR for 3D lane detection using transformers and achieved an F1 score of 80.59\% \cite{luo2023latr}. Che et al. developed TwinLiteNet and later TwinLiteNetPlus  to deliver lightweight segmentation with lane IoU scores of 31.08\% and 34.2\%, respectively \cite{che2023twinlitenet}\cite{che2024twinlitenetplus}. Similarly, models like CurveFormer++ \cite{bai2025curveformer++}, SRLane \cite{chen2024sketch}, and ENet-21 \cite{hosseini2024enet} have improved lane detection using various CNN and attention-based architectures, with F1 scores surpassing 77\%.

 Substantial research is reproted for traffic sign recognition. Dang et al. Used YOLOv5 for real-time TSR in adverse conditions, reaching 79.2\% accuracy \cite{dang2023improved}. Dalborgo et al. adapted YOLOv5 for Brazilian vegetation-occluded signs with 95.1\% precision \cite{dalborgo2023traffic}, while Han et al. introduced YOLO-SG and achieved over 97\% accuracy for danger signs \cite{han2024yolo}. Other notable contributions include haze-resistant TSR by Radha Rani et al. \cite{rani2024traffic}, YOLO-TS with improved receptive fields \cite{chen2024yolo}, EMDFNet with 97.0\% mAP on GTSDB \cite{li2024emdfnet}, and YOLO-PPA, optimized for cruise control applications \cite{zhang2024yolo}.

To handle multiple ADAS tasks simultaneously, recent models have adopted multi-task learning strategies. Key manifestations are observed in YOLOP \cite{wu2022yolop}, HybridNets \cite{vu2022hybridnets}, and YOLOPv2 \cite{han2022yolopv2}, these efforts integrate object detection, lane marking, and drivable area segmentation in real-time. Ye and Zhang proposed Mobip using MobileNetV2 \cite{ye2023mobip}, while CenterPNets \cite{chen2023centerpnets}, YOLOM \cite{wang2024you}, and YOLOP-DN \cite{tan2024multi} optimized these systems further for  deployment on embedded systems. The latest advancements such as YOLOPv3, YOLOPX, and BiFPN-based models have pushed lane accuracy up to 88.6\% with improved efficiency \cite{zhan2024multi} \cite{batista2023application} \cite{zhang2024high}.

However, despite these global advancements, the variability in lane markings, occlusions due to traffic or debris, adverse weather conditions, and non-standardized traffic signs pose unique challenges. While lane detection models benefit from relatively standardized road structures, traffic sign recognition faces greater diversity across countries, particularly in developing regions where standardized datasets are scarce. Most existing research relies on publicly available datasets from European and developed countries, such as the German Traffic Sign Detection Benchmark (GTSDB), Belgium Traffic Sign Dataset (BTSD), LISA (USA), Italian Traffic Signs Dataset (DITS), Swedish Traffic Signs Dataset (STSD) etc \cite{triki2024comprehensive},\cite{flores2024traffic}.Although these benchmarks have driven considerable progress, their emphasis on structured and well-annotated environments limits their applicability to real-world conditions in third-world countries.

In Pakistan, very limited research has been done in domain of Trafic sign detection and reconition. The complexity of local traffic scenes and the lack of robust models tailored to such environments reflect the low recognition accuracies reported by existing studies on Pakistani traffic sign datasets, which range from approximately 41\% \cite{nadeem2018transfer}, 55\% \cite{nadeem2022pakistani}, to at best 83.4\% \cite{sikander2021image}. This clear performance gap and scarcity of annotated regional datasets highlight the need for robust and resource-efficient TSR solutions capable of adapting to diverse, low-resource, and non-standardized traffic conditions.

The main contributions of this work are summarized as follows:
\begin{itemize}
\item Propose a lightweight, edge-deployable ADAS framework that integrates lane departure warning and traffic sign recognition in real time.
\item Design and implementation of a custom traffic sign recognition pipeline using YOLO-based detection and a lightweight classification model optimized for local road conditions.
\item Validation of the traffic sign classification model's cross dataset generalization by fine tuning it on the Bangladesh traffic sign dataset and German Traffic Sign Recognition Benchmark using the same augmentation profile and second stage training strategy.
\item Development of a lane detection and segmentation approach using TwinLiteNet, followed by a custom-designed lane departure warning algorithm based on geometric distance analysis.
\item Real-time implementation and evaluation of the complete framework on edge devices, demonstrating its practicality for resource constrained environments.
\end{itemize}

The rest of this paper is organized as follows: Section II introduces the proposed framework. Section III presents traffic sign detection and recognition, including dataset collection, augmentation, and model training. Section IV discusses lane detection, lane departure warning, and the fallback mechanism. Section V describes the integration of lane departure warning with traffic sign recognition into a unified system. Section VI highlights the real-time implementation of the framework on edge devices. Section VII outlines future research directions.

\section{Proposed Framework}

\begin{figure}
    \centering
    \includegraphics[width=1\linewidth, keepaspectratio]{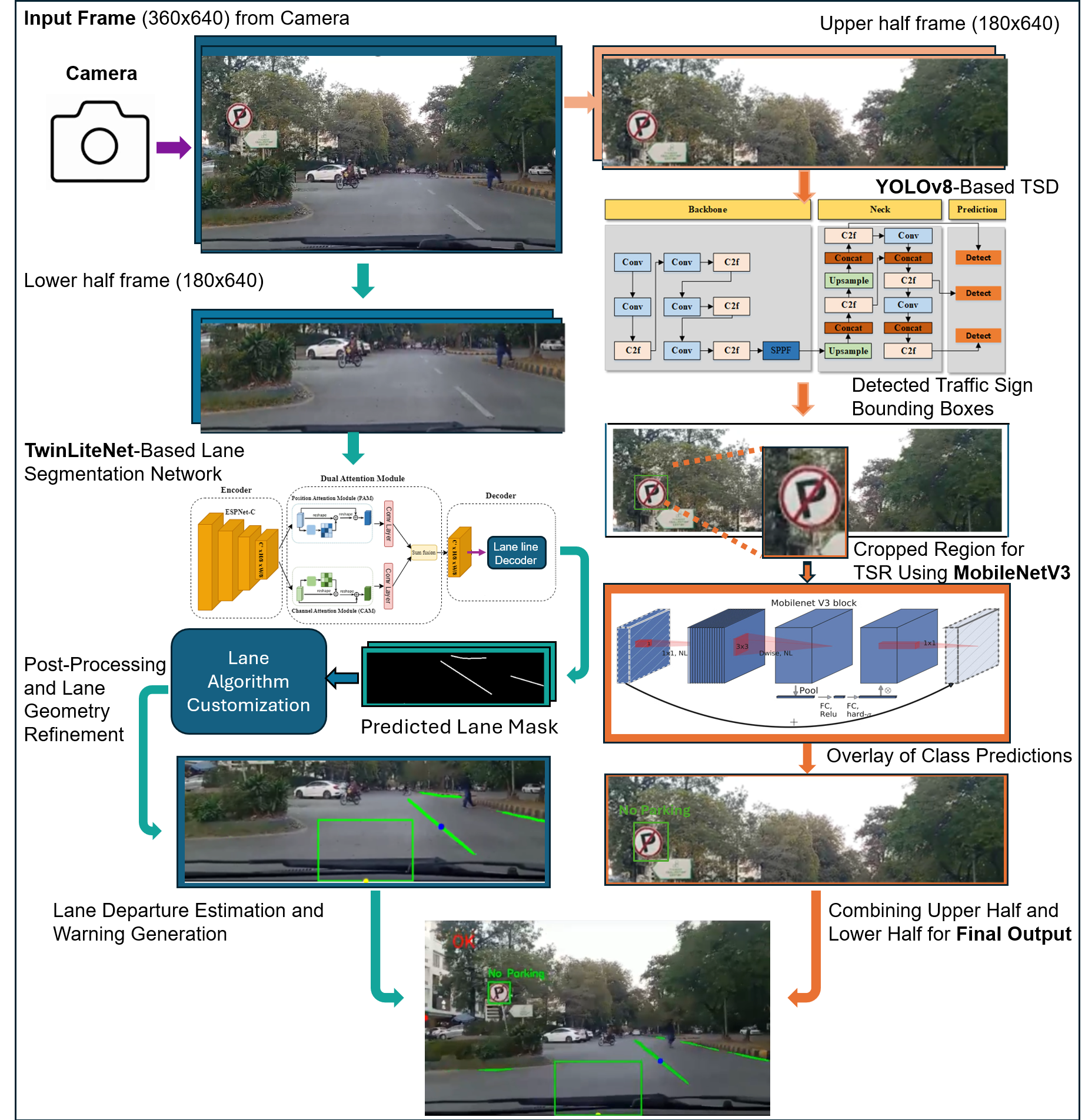}
    \caption{Overall Framework of the Proposed Intelligent Driving Assistance System. YOLOv8s\cite{yaseen2024yolov9} for traffic sign detection, MobileNetV3-small \cite{howard2019searching} for traffic sign recognition and TwinLiteNet  \cite{che2023twinlitenet} for lane detection }
    \label{fig:overview}
\end{figure}

The framework of the proposed intelligent driving assistance system is designed to be modular for real-time performance on embedded platforms. The framework integrates three deep learning models: TwinLiteNet  \cite{che2023twinlitenet} for lane detection, YOLOv8s\cite{yaseen2024yolov9} for traffic sign detection, and MobileNetV3-small \cite{howard2019searching} for traffic sign recognition. Complete Framework  shown in Figure~\ref{fig:overview}.

The real-time perception process begins with system initialization, in which all models are loaded into the device's memory (Jetson Orin Nano). Simultaneously, the onboard camera captures a live stream from the vehicle's viewpoint. Each frame is resized to a consistent resolution of $640 \times 360$ pixels and divided horizontally into two equal parts:
\begin{itemize}
    \item The upper half ($640 \times 180$) is dedicated to traffic sign analysis.
    \item The lower half ($640 \times 180$) is reserved for lane analysis.
\end{itemize}
This logical division supports a pipelined processing approach where traffic sign analysis is performed firstly and that is followed by lane analysis. This approach reduce the computational cost for each model and accelerate overall inference.

In the upper half of the image, the YOLOv8s model performs object detection to locate traffic signs. The cropped regions of interest (ROIs) containing detected signs are then passed to the MobileNetV3-small classifier, which accurately identifies the category of each sign, such as speed limits, U-Turn, or No Parking.

For lightweight and efficient classification, MobileNetV3 integrates three key components: depthwise separable convolutions, squeeze-and-excitation (SE) modules, and hard-swish activation function \cite{howard2019searching}. Among these, the hard-swish non-linearity is mathematically expressed as:

\[
\scalebox{1}{$
\text{h-swish}(x) = x \cdot \frac{\text{ReLU6}(x+3)}{6}.
$}
\]
This activation improves accuracy compared to ReLU and it is efficient for edge deployment. The MobileNetV3 is well suited for real-time recognition tasks on embedded platforms.

Concurrently, the lower half of the frame is processed by TwinLiteNet, which is a lightweight dual-decoder network well suited for lane detection and drivable area segmentation \cite{che2023twinlitenet}. The encoder is based on ESPNet-C, which efficiently extracts multi-scale feature representations. To further enhance contextual learning, a Dual Attention Module is employed, which is composed of the Position Attention Module (PAM) for capturing spatial dependencies and the Channel Attention Module (CAM) for modeling inter-channel relationships. Their outputs are fused via convolution and element-wise summation to form the feature map \[
B \in \mathbb{R}^{32 \times \frac{H}{8} \times \frac{W}{8}}
\]

TwinLiteNet uses a multi output architecture with two lightweight decoders, one for drivable area segmentation and the other for lane detection. The separation allows each task to be improved at a low cost. Both Tversky Loss and Focal Loss are combined into a loss function. Focal Loss is defined as

\[
\scalebox{1}{$
\mathcal{L}_{focal} = -\frac{1}{N}\sum_{c=0}^{C-1}\sum_{i=1}^{N} p_i(c) (1 - \hat{p}_i(c))^\gamma \log(\hat{p}_i(c))
$}
\]
where $N$ is the total number of pixels, $C$ is the number of classes, $p_i(c)$ is the ground-truth label for pixel $i$ of class $c$, $\hat{p}_i(c)$ is the predicted probability, and $\gamma$ is a focusing parameter that emphasizes hard-to-classify pixels. This loss mitigates class imbalance by down-weighting well-classified pixels. Tversky Loss is defined as

\[
\scalebox{1}{$
\mathcal{L}_{tversky} =
\sum_{c=0}^{C}\!\left(1-\frac{TP(c)}{TP(c)+\alpha FN(c)+\beta FP(c)}\right)
$}
\]

where $TP(c)$, $FN(c)$, and $FP(c)$ denote the true positives, false negatives, and false positives for class $c$, and $\alpha, \beta$ control the relative importance of false negatives and false positives. This loss allows the network to focus on accurate detection of sparse or difficult features such as lane markings. The total training objective combines both losses:
\[
\scalebox{1}{$
\mathcal{L}_{total} = \mathcal{L}_{focal} + \mathcal{L}_{tversky}.
$}
\]
During inference, convolution and batch normalization layers are re-parameterized into a single equivalent convolution to reduce latency with a high accuracy. The network produces a pixel-wise lane mask with the detected lane boundaries, which is then analyzed by a custom lane departure detection algorithm to estimate the vehicle's lateral deviation relative to the lane center and issue warnings when departures are detected.

Finally, the outputs from both tasks are merged and overlaid on the original frame. The top section visualizes recognized traffic signs, while the bottom section displays lane markings and any active departure warnings. The designed framework ensures scalability and supports deployment on edge devices for autonomous vehicles. 

In the following sections, we will provide  details, real-time implementation, methodologies, and performance outcomes of each module, i.e. traffic sign detection and recognition, lane segmentation and departure analysis. 

\section{Traffic Sign Detection \& Recognition}

\begin{figure}
    \centering
    \includegraphics[width=0.8\linewidth]{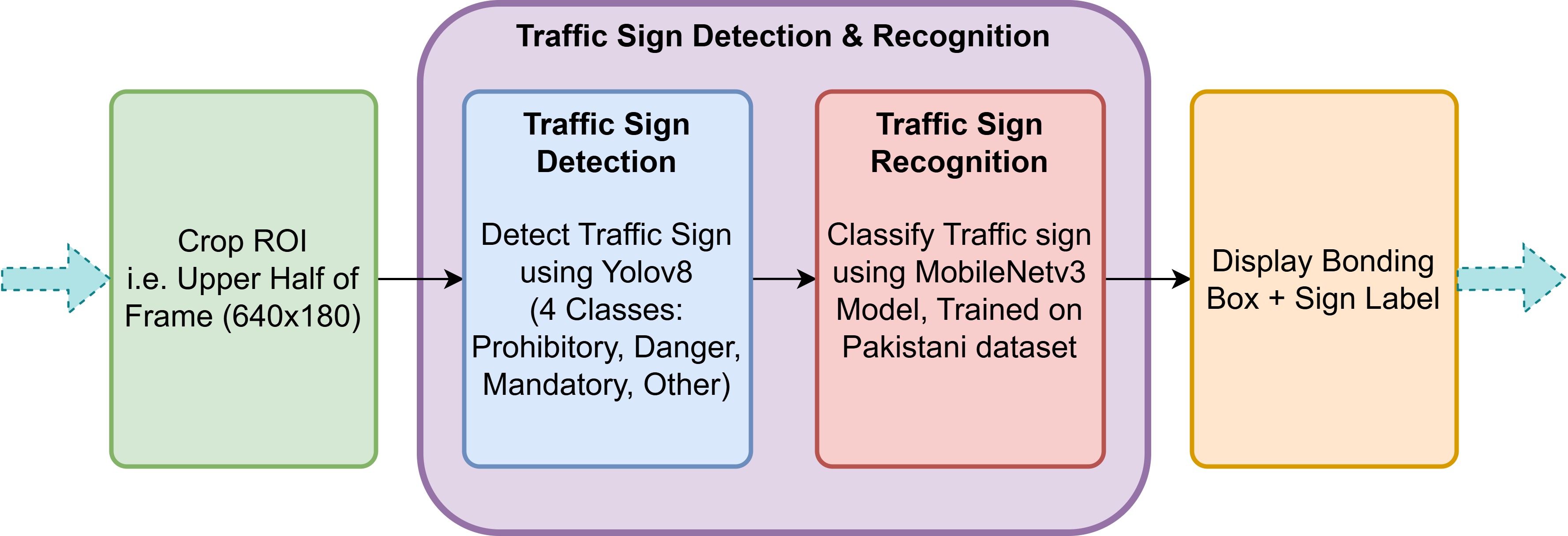}
    \caption{Traffic Sign Detection and Recognition Flow}
    \label{fig:TSDR}
\end{figure}

The detection and identification of traffic signs is essential for driver's assistance and facilitating the operations of semi-autonomous vehicles. Such technologies ensure adherence to traffic regulations and enhance road safety by precisely recognizing warning and regulatory signals. Detection and recognition are the two primary phases of the traffic sign processing pipeline (as illustrated in Figure~\ref{fig:TSDR}). Roadside signs are usually visible in the upper half of each input frame (640 × 180 resolution), which is processed by the system. Traffic signs are detected and classified into four main categories in the first stage using the YOLOv8s model: Prohibitory, Mandatory, Danger, and Other.

Although YOLO's one-stage architecture \cite{redmon2016you} is highly efficient and achieves real-time performance, its shared detection–classification framework often prioritizes speed at the expense of accuracy. As a result, small or distant objects are frequently misclassified \cite{olorunshola2023comparative}, \cite{youssouf2022traffic},\cite{safdar2025yoloatr}. In contrast, two-stage detectors such as Faster R-CNN and Mask R-CNN generally provide higher accuracy by decoupling detection and classification, but this improvement comes with significantly slower inference speed \cite{zhou2025multi}, \cite{olorunshola2023comparative}.

To overcome this challenge, we propose a two-stage pipeline with a lightweight classifier. In the first stage, YOLOv8s is used solely for localizing traffic signs and classify them under any of the four broad categories. These category assignments are not critical in themselves even if YOLO introduces some mis-classification at this stage. The essential requirement is accurate detection of the sign's location. The second stage performs the detailed recognition task using the MobileNetV3-small model, a compact CNN optimized for image classification \cite{howard2019searching}. By processing higher-resolution cropped traffic sign patches, the classifier ensures accurate categorization of the 35 distinct traffic sign classes (listed in Table S1 of Supplementry Information) \cite{mexwell_pakistanitrafficsign_2025}. The integration of YOLOv8s for detection and MobileNetV3-small for classification thus combines real-time efficiency with fine-grained recognition for reliable performance in intelligent transportation applications. It is appropriate for real-time applications in local traffic scenarios because this two-stage pipeline guarantees both effective detection and high-accuracy identification.

\subsection{Traffic Sign Detection}
In order to identify the most efficient model for real-time Traffic Sign Recognition in Pakistan, several deep learning architectures were evaluated. The Intel Core i5 8th Gen is the system that initial assessments were conducted on for baseline performance. FR-CNN's slow inference speed made it unsuitable for dynamic driving. While lightweight models like YOLOv8n offered high frames-per-second, they exhibited a lower Recall of 0.8111, increasing the risk of missing critical signs in the high-variability environments common in underdeveloped regions.

YOLOv8s was selected as the optimal model as demonstrated in Table \ref{tab:model_comparison}. It achieved a superior balance of safety and performance, yielding the highest mAP50 (0.9707) and Recall (0.9092), outperforming even the larger YOLOv8m variant. YOLOv8s ensures both accuracy and efficiency by achieving a real-time capable speed of 3.63 frames per second during initial tests. A robust solution tailored to the specific linguistic and structural complexity of Pakistani road signs is provided by this model.

\begin{table}[!ht]
\centering
\caption{Comparison of Different Models for Traffic Sign Detection (Tested on Dell Latitude, Core i5 8th Gen with 16 GB RAM and 3.6 GHz processor without GPU).}
\label{tab:model_comparison}
\setlength\tabcolsep{2pt}
\begin{tabular}{l c c c c c c c }
\toprule
\textbf{Model} & \textbf{Params} & \textbf{Framework} & \textbf{Precision} & \textbf{Recall} & \textbf{mAP50} & \textbf{mAP50-95} & \textbf{FPS} \\
\midrule
\multirow{2}{*}{FR-CNN} & \multirow{2}{*}{42M} & PyTorch & \multirow{2}{*}{-} & \multirow{2}{*}{-} & \multirow{2}{*}{0.595} & \multirow{2}{*}{0.324} & 0.30 \\
& & ONNX & & & & & 0.59 \\
YOLOv5s & 7.2M & PyTorch & 0.8977 & 0.7970 & 0.8665 & 0.5958 & 5.40 \\
YOLOv8n & 3.2M & PyTorch & 0.9826 & 0.8111 & 0.9025 & 0.7192 & 7.50 \\
\textbf{YOLOv8s} & \textbf{11.2M} & \textbf{PyTorch} & \textbf{0.9740} & \textbf{0.9092} & \textbf{0.9707} & \textbf{0.7996} & \textbf{3.63} \\
YOLOv8m & 25.9M & PyTorch & 0.9893 & 0.8935 & 0.9540 & 0.7964 & 1.32 \\
\bottomrule
\end{tabular}%
\end{table}

\subsection{Traffic Sign Recognition}
 In this work, we focus is on recognizing specific traffic signs in Pakistan to align the model with local road conditions and regulations. This research required a specialized dataset representing the unique traffic sign designs and symbols used in Pakistan. unlike generic datasets that contain a mix of international traffic signs. When deployed in real world scenarios, this customization was necessary to ensure the system's reliability and accuracy.

\subsubsection{Dataset Collection and Augmentation}

\begin{figure}
    \centering
    \includegraphics[width=1\linewidth]{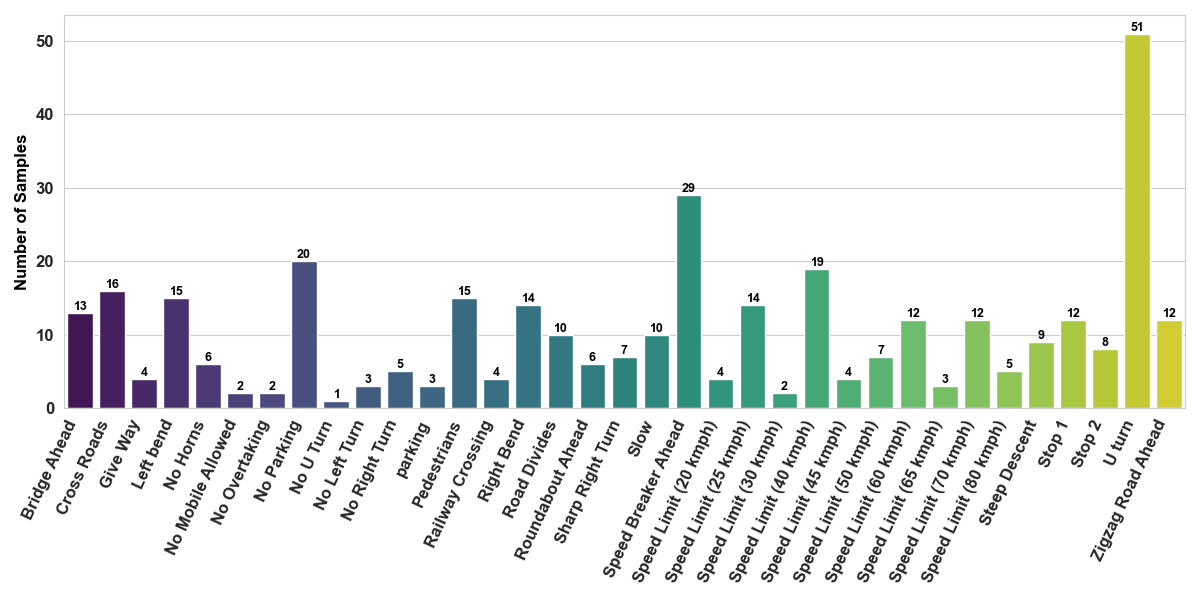}
    \caption{Dataset obtained from Kaggle \cite{mexwell_pakistanitrafficsign_2025} containing 359 samples}
    \label{fig:kaggle_dataset}
\end{figure}
\begin{figure}
    \centering
    \includegraphics[width=0.8\linewidth]{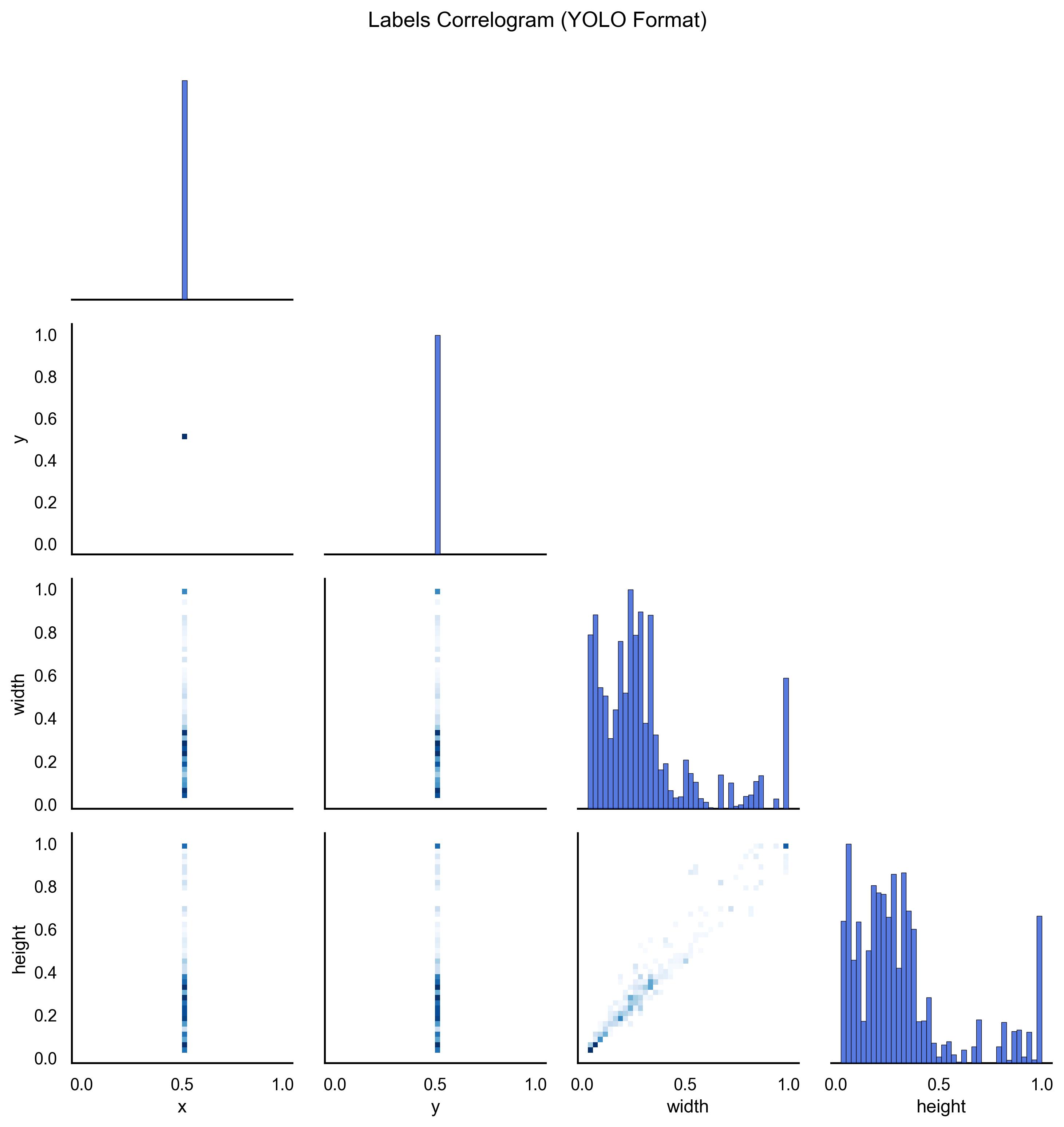}
    \caption{Labels correlogram showing spatial distributions of traffic sign annotations. The concentration of $x$ and $y$ coordinates at the center reflects the alignment achieved after targeted cropping and augmentation.}
    \label{fig:labels_correlogram}
\end{figure}

Pakistani traffic signs dataset covering 35 classes (listed in Table S1 of Supplementry Information) were obtained from Kaggle  \cite{mexwell_pakistanitrafficsign_2025}. The risk of overfitting and poor generalization to real world scenarios was high due to the limited size of the dataset with only total of 359 samples as shown in Figure~\ref{fig:kaggle_dataset}. The Roboflow platform \cite{roboflow2025} is an intuitive tool for dataset preparation and augmentation. An augmentation profile was designed to mimic real world variations, which included random blur, viewpoint shifts (tilting), adjustments in brightness and both high/low saturation levels. The dataset was expanded to 3,764 images by these augmentations.

The persistence of underrepresented classes was addressed by a second round of targeted augmentation. An augmentation profile that included rotation, blur, and adjustments to saturation and exposure was applied in this stage. To enhance the model's robustness to extreme illumination changes, a particular emphasis was placed on brightness-based transformations. A uniform distribution of 250 samples per class was ensured as a result of this strategic expansion. The geometric consistency of this augmented dataset improved the model's ability to generalize across diverse real-world environmental variations, as shown in Figure~\ref{fig:labels_correlogram}. The distribution of normalized bounding box parameters ($x, y, \text{width}, \text{height}$) shows that the  $x$ and $y$ coordinates are tightly concentrated around the image center. The width and height distributions are skewed towards smaller, uniform values with a positive correlation between them suggesting stable aspect ratios across all categories. Ultimately, the correlogram confirms that the dataset maintains uniform annotation characteristics and consistent object alignment, providing a reliable foundation for model training.

\subsubsection{Model Training on Custom Dataset}
Initially, a pretrained ReXNet\_150 model was tested for traffic sign classification. However, due to its large weight size of around 9.8 million parameters, the model achieved an inference speed of only 6 FPS, which is not suitable for real time applications. Furthermore, since YOLOv8s (used for detection) is itself a relatively heavy model, combining it with ReXNet\_150 further reduced the speed to only 1.5 to 2 FPS on the local machine (Intel Core i5 laptop without GPU), making the pipeline impractical for real-time deployment. 


\begin{table}[!ht]
\centering
\small
\caption{Hyperparameters for MobileNetV3-small}
\label{tab:mobilenet_training_settings_1}
\begin{tabular}{l p{4.5cm}}
\toprule
\textbf{Parameter} & \textbf{Value} \\
\midrule
Model Architecture      & MobileNetV3-small \\
Parameters              & 2.54 Million \\
Dataset Size            & 3764 Images \\
Dataset Split           & 65\% Training, 15\% Validation, 20\% Testing \\
Learning Rate           & 0.0005 \\
Batch Size              & 32 \\
No. of Epochs           & 50 \\
Stopping Criteria       & 50 epochs \textit{or} Validation Loss not decreasing for 5 epochs \\
Framework Used          & PyTorch \\
\bottomrule
\end{tabular}
\end{table}

\begin{table}[!ht]
\centering
\small
\caption{Training Results for MobileNetV3-small}
\label{tab:mobilenet_training_results_1}
\begin{tabular}{l p{3.5cm}}
\toprule
\textbf{Parameter} & \textbf{Value} \\
\midrule
Model Accuracy on Validation Set        & 94.4\% \\
Model Accuracy on Test Set              & 92.6\% \\
No. of Epochs in Training               & 46 \\
No. of Images in Inference Set (Self Collected) & 58 \\
Correctly Classified                    & 34 \\
Misclassified                           & 24 \\
Accuracy on Inference Set               & 58.5\% \\
\bottomrule
\end{tabular}
\end{table}

\begin{figure}
    \centering
    \includegraphics[width=0.5\linewidth]{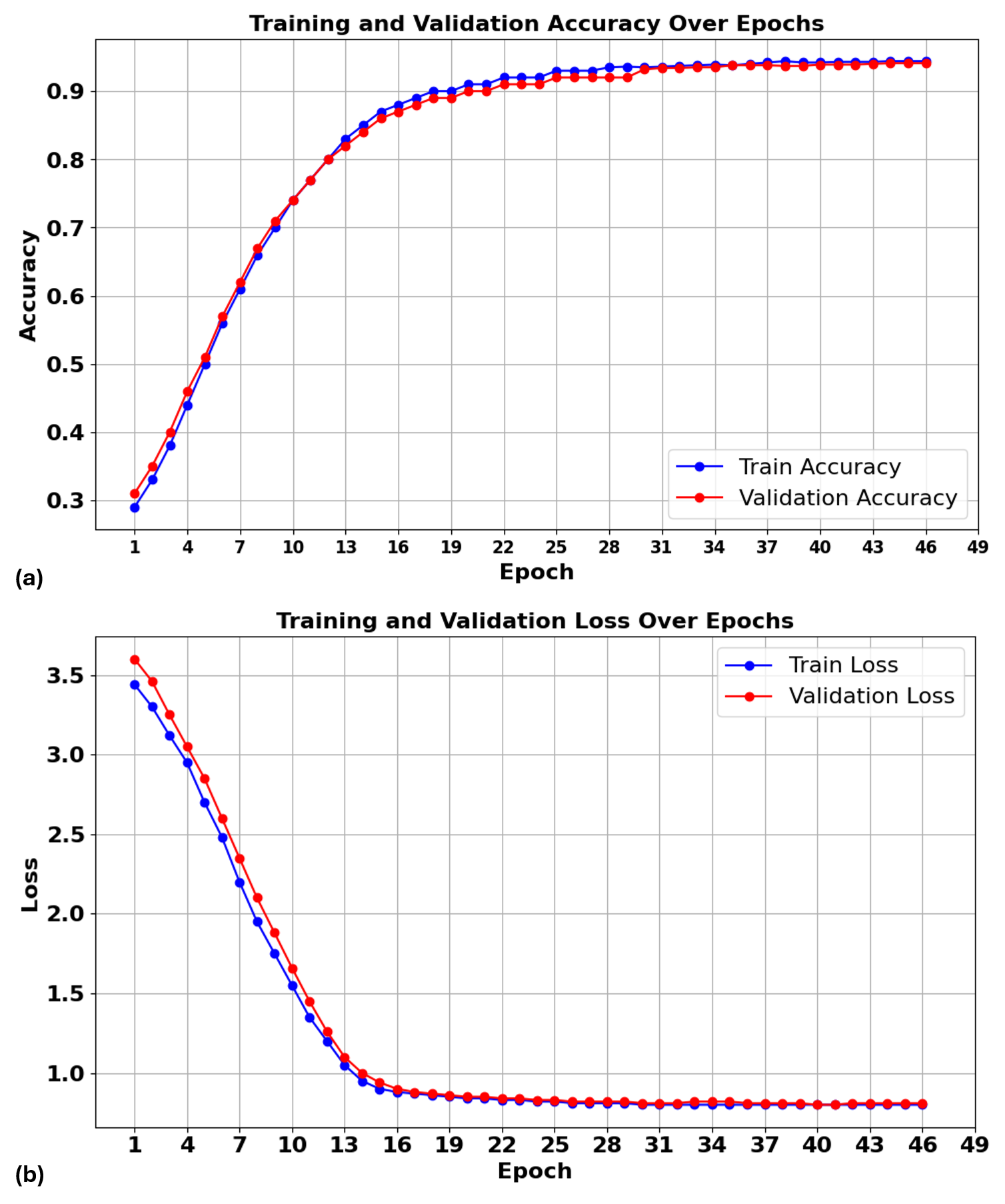}
    \caption{(a) Training and Validation Accuracy (b) Training and Validation Loss}
    \label{fig:training 1}
\end{figure}

To address this limitation, MobileNetV3-small a lightweight model with only 2.54 million parameters was adopted due to its efficiency on edge devices. The model was trained on an NVIDIA Tesla P100 GPU (16 GB HBM2 memory, 3,584 CUDA cores), achieving an average training throughput of roughly 230 images per second (approx 7.2 iterations/s × 32 images per batch). The initial training of MobileNetV3-small on an imbalanced dataset of 3,764 images (Hyperparameters shown in Table~\ref{tab:mobilenet_training_settings_1} and Training graph shown in Figure~\ref{fig:training 1}) achieved 92.6\% of test set accuracy but on real world samples (Self collected) correctly identifying only approximately 58\%(Training results shown in Table~\ref{tab:mobilenet_training_results_1}). This result highlighted the need for a more balanced dataset to improve generalization.

\begin{table}[!ht]
\centering
\small
\caption{Hyperparameters of 2nd Stage Training for MobileNetV3-small}
\label{tab:mobilenet_training_settings}
\begin{tabular}{l p{4.5cm}}
\toprule
\textbf{Parameter} & \textbf{Value} \\
\midrule
Model Architecture      & MobileNetV3-small \\
Parameters              & 2.54 Million \\
Dataset Size            & 8750 Images \\
Dataset Split           & 70\% Training, 20\% Validation, 10\% Testing \\
Learning Rate           & 0.0005 \\
Batch Size              & 32 \\
No. of Epochs           & 50 \\
Stopping Criteria       & 50 epochs \textit{or} Validation Loss not decreasing for 5 epochs \\
Framework Used          & PyTorch \\
\bottomrule
\end{tabular}
\end{table}

\begin{table}[!ht]
\centering
\small
\caption{2nd Stage Training Results for MobileNetV3-small}
\label{tab:mobilenet_training_results_2}
\begin{tabular}{l p{3.5cm}}
\toprule
\textbf{Parameter} & \textbf{Value} \\
\midrule
Model Accuracy on Validation Set        & 95.6\% \\
Model Accuracy on Test Set              & 97.83\% \\
No. of Epochs in Training               & 18 \\
No. of Images in Inference Set (Self Collected) & 58 \\
Correctly Classified                    & 54 \\
Misclassified                           & 4 \\
Accuracy on Inference Set               & 93.1\% \\
\bottomrule
\end{tabular}
\end{table}

\begin{figure}
    \centering
    \includegraphics[width=0.5\linewidth]{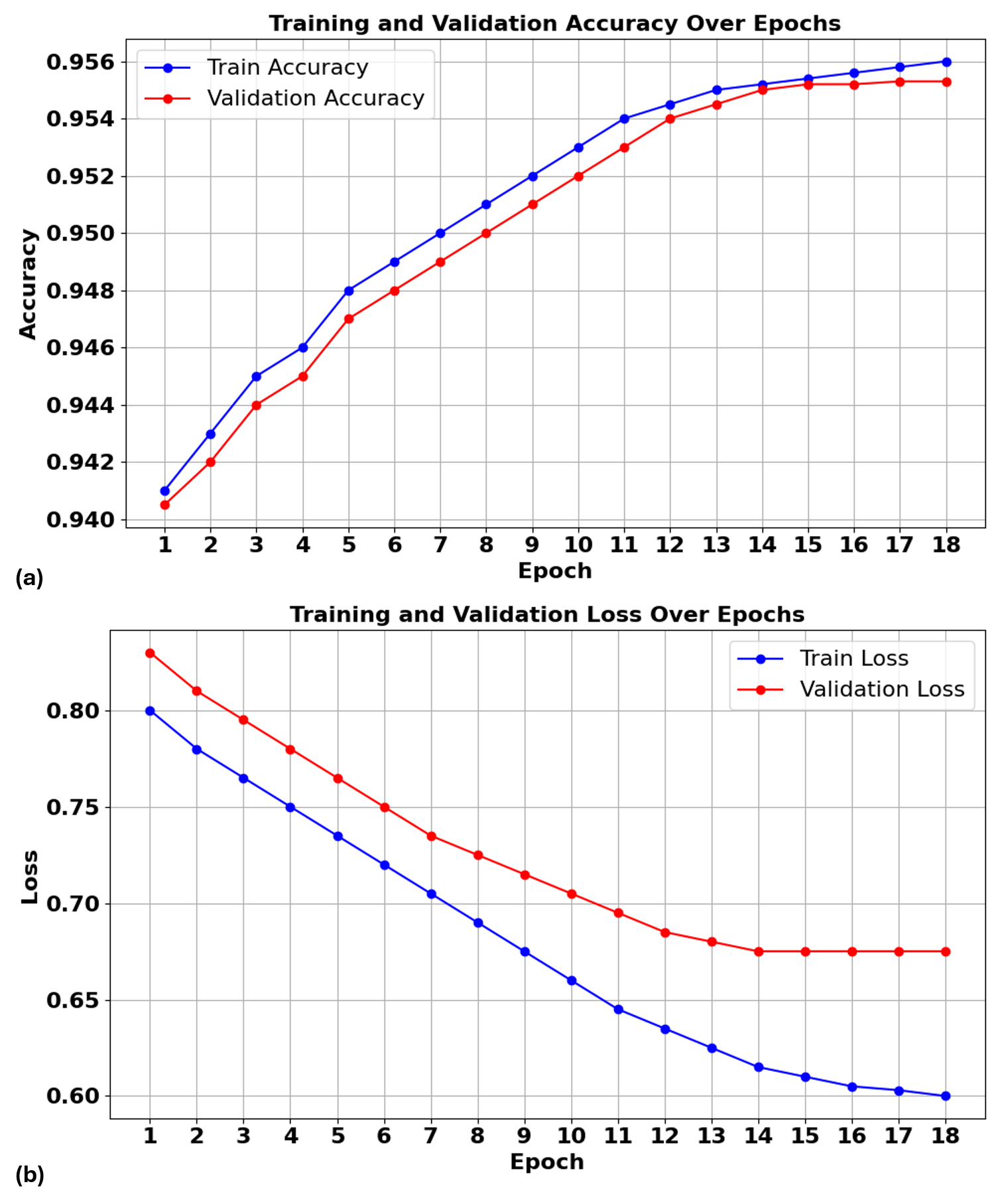}
    \caption{(a) 2nd Stage Training and Validation Accuracy (b) 2nd Stage Training and Validation Loss}
    \label{fig:training 2}
\end{figure}

\begin{figure}
    \centering
    \includegraphics[width=0.9\linewidth]{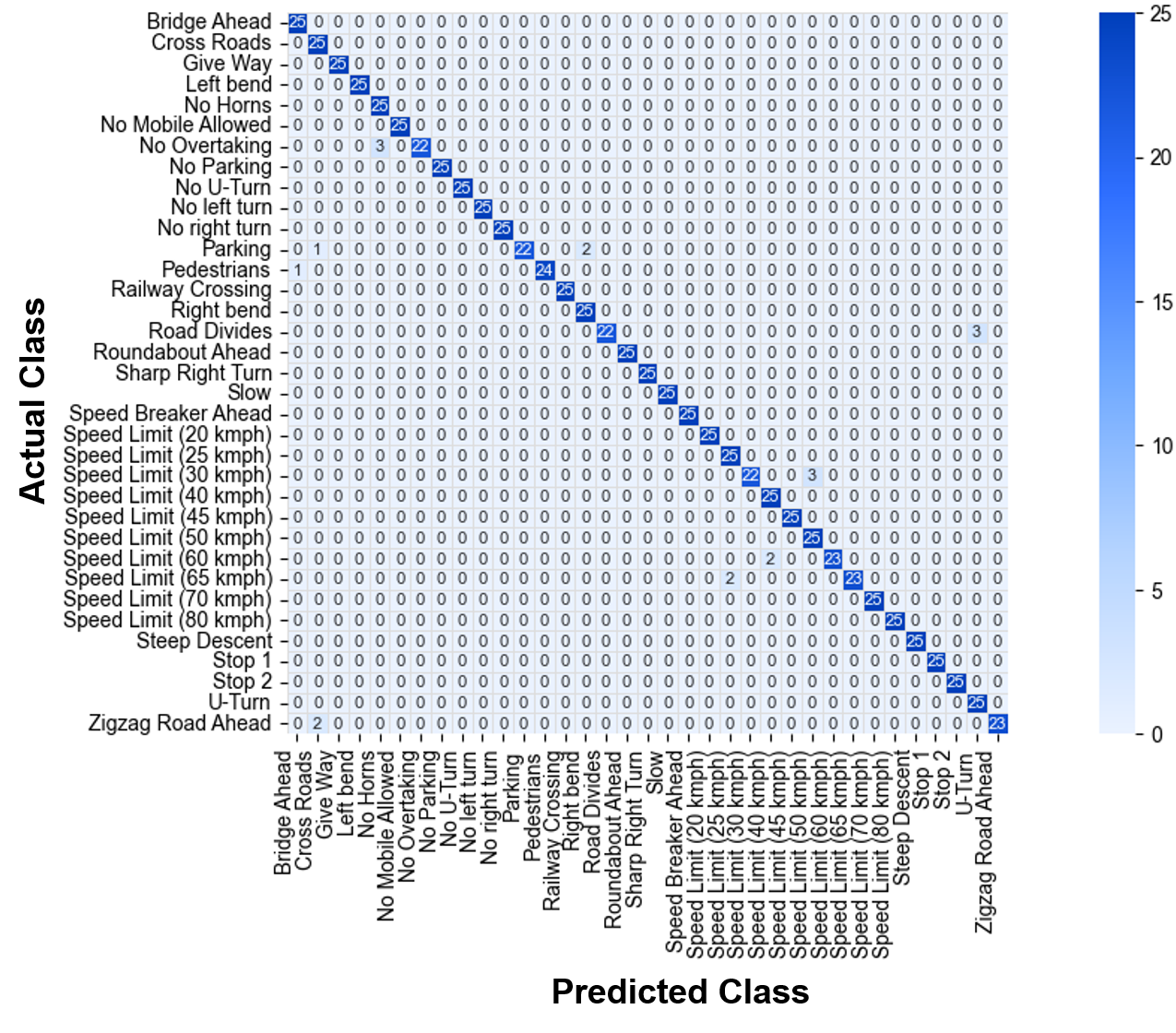}
    \caption{Confusion Matrix of  MobileNetV3-small on Test set }
    \label{fig:confusion}
\end{figure}

Therefore, a second-stage training phase was conducted using a balanced dataset of 8,750 images. The training configuration was kept the same, except for a slight change in the split percentage as shown in Table~\ref{tab:mobilenet_training_settings}. Training Graph shown in Figure~\ref{fig:training 2}. As a result, the MobileNetV3-small model achieved test set accuracy of 97.13\% (Confusion Matrix Shown in Figure~\ref{fig:confusion}) and a significantly improved accuracy of 93.1\% on real world samples (Results of Second Stage Training shown in Table~\ref{tab:mobilenet_training_results_2}). Moreover, inference speed tests on a standard Intel Core i5 laptop (without GPU) demonstrated that MobileNetV3-small processed each frame in approximately 12 ms (Approx 83 FPS), vastly outperforming ReXNet\_150 which required around 167 ms per frame (Approx 6 FPS). This combination of accuracy and efficiency established MobileNetV3-small as the ideal solution for real-time deployment of traffic sign recognition in Pakistani driving environments.

\textbf{Cross-Dataset Generalization Validation}
To evaluate the model's ability to generalize beyond the local dataset, the base model originally trained on the Pakistani traffic sign dataset was fine-tuned and validated across diverse geographic and structural contexts. First, the model was fine-tuned on the Bangladesh traffic sign dataset \cite{tusher2023comparative}, \cite{Tusher2020TrafficSignBangladesh} using the same augmentation profile (brightness, saturation, exposure, rotation, and blur) and the second stage training configuration detailed in Table~\ref{tab:mobilenet_training_settings}. The model achieved a validation accuracy of 96.6\% and a test accuracy of 95.7\% on this dataset. In third-world environments, where traffic signs face challenges such as inconsistent maintenance, non-standard positioning, and varying environmental degradation, the system's reliability is underscored by these results.

To further assess the model scalability against large scale benchmark, The German Traffic Sign Recognition Benchmark (GTSRB) dataset \cite{Houben-IJCNN-2013} was also used to fine-tune the base variant of our model. The model achieved a solid test accuracy of 88.4\% despite the significant increase in class complexity and environmental variance inherent in this globally recognized dataset. By maintaining high performance across both specialized regional datasets and extensive wellknown benchmarks, the proposed model demonstrates robust generalization capabilities. The architecture is not only effective for the unique challenges of developing nations, but it is also capable of competitive performance on larger, standard datasets used in global self driving research.

\subsection{Integration of Traffic Sign Detection and Recognition}
\begin{figure}
    \centering
    \includegraphics[width=1\linewidth]{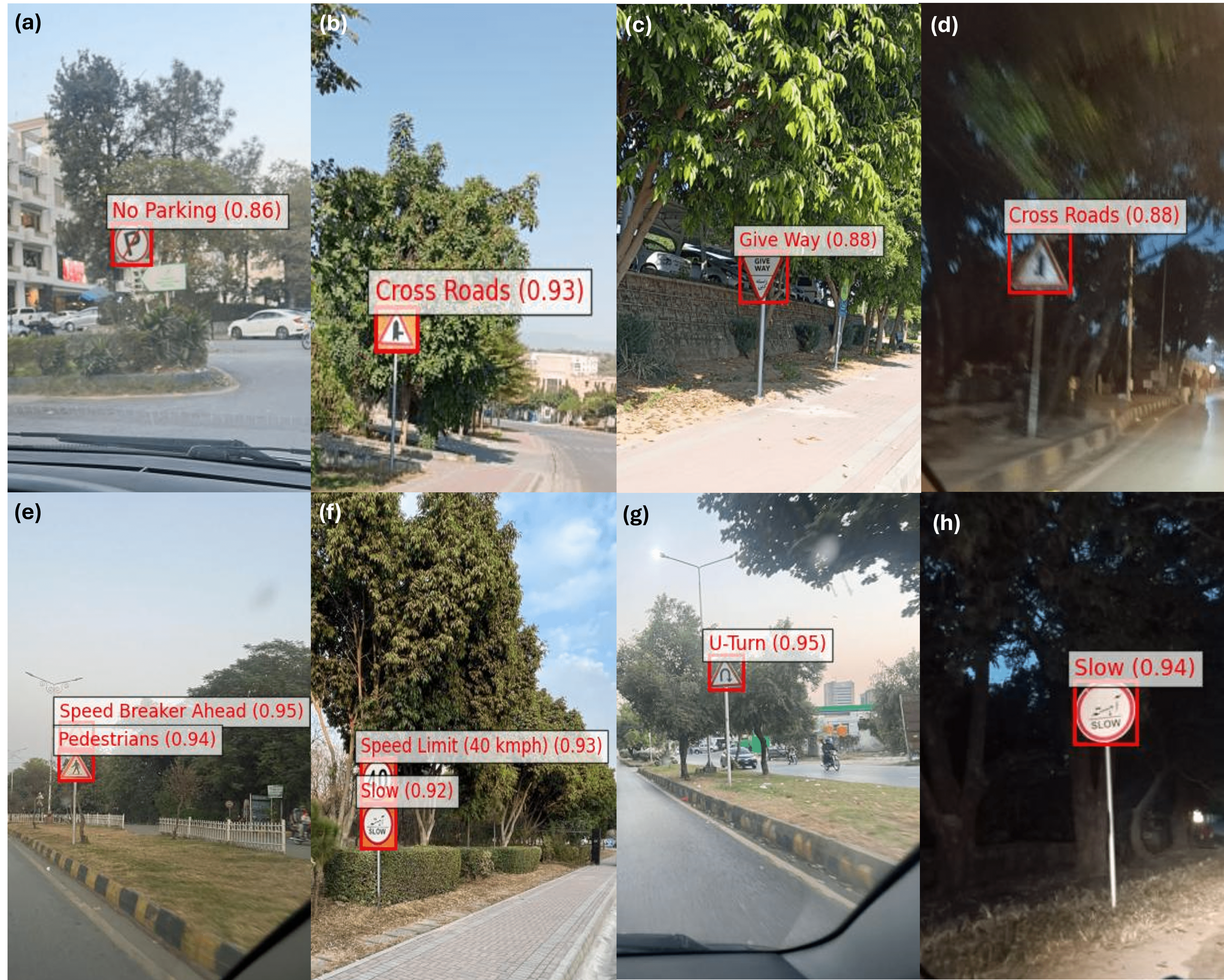}
    \caption{Detecting and Classifying Traffic Signs. The traffic sign detected and classified in image (a) No Parking, (b) Cross Roads, (c) Give way, (d) Cross Roads, (e) Speed Breaker and Pedestrian, (f) Speed Limit (40 Kmph) and Slow, (g) U-Turn and (h) Slow}
    \label{fig:TSD&R}
\end{figure}

Real-time and accurate recognition is achieved by the traffic sign identification system. The YOLOv8s model can detect traffic signs within an input frame by localizing and drawing bounding boxes around relevant sign regions. The detected regions are passed to a MobileNetV3-small classifier, which performs detailed recognition by accurately identifying specific traffic sign types from the cropped images. Efficient processing and reliable performance in traffic environments are ensured by this division of tasks detection. Results shown in Figure~\ref{fig:TSD&R}. There is a detailed comparison between the proposed framework and existing traffic sign detection and recognition approaches shown in Table~\ref{tab:comparison_matrix}. Early studies relied on lightweight CNNs such as LeNet-5 and reported limited recognition performance, reflecting the complexity of local traffic scenes and lack of robust models tailored to developing-country conditions. The proposed framework achieved a significantly higher 93.1\% accuracy and 97.4\% mAP on the Pakistani dataset, while maintaining real-time performance of 18FPS on an NVIDIA Orin Nano, demonstrating clear improvement under practical edge-deployment constraints. Without reporting real-time performance, prior MobileNetV2 and VGG16-based approaches achieved 89.65\% and 90.31\% accuracy for the Bangladeshi traffic sign dataset. The proposed method further improves recognition accuracy to 95.7\%, showing its strong generalization across traffic environments in developing regions. Although several recent detection-oriented methods such as EMDFNet and YOLO-PPA report high mAP values on benchmark datasets,but these are not validated on edge device. The results show that the proposed two-stage YOLOv8s and MobileNetV3-small framework offers a favorable trade-off between accuracy, computational efficiency, and deployability. The proposed approach provides a practical andScalable solution for realworld ADAS applications in developing regions by achieving superior performance on Pakistani and Bangladeshi datasets while running on resourceconstrained edge hardware.

\begin{table}[]
\centering
\caption{Comparison of Different Models for Traffic Sign Detection and Recognition}
\label{tab:comparison_matrix}
\resizebox{\linewidth}{!}{%
\begin{tabular}{l c c c c c c c}
\toprule
\textbf{Model} & \textbf{Year} & \textbf{mAP (\%)} & \textbf{Accuracy (\%)} & \textbf{Params (M)} & \textbf{FPS} & \textbf{Device} & \textbf{Dataset} \\
\midrule

Pakistani TSR (LeNet-5) \cite{nadeem2018transfer} & 2018 & -- & 41.0 & 0.062 & -- & -- & Pakistani TSR \\ \midrule

Pakistani TSR (LeNet-5) \cite{sikander2021image} & 2021 & -- & 83.18 & 0.062 & -- & -- & Pakistani TSR \\ \midrule

Pakistani TSR (Transfer Learning) \cite{nadeem2022pakistani} & 2022 & -- & 55.0 & 0.062 & -- & -- & Pakistani TSR \\ \midrule

Bangladesh TSR (MobileNetV2) \cite{Tusher2020TrafficSignBangladesh} & 2023 & -- & 89.65 & 3.4 & -- & -- & Bangladesh Dataset \\ \midrule

Bangladesh TSR (VGG16) \cite{Tusher2020TrafficSignBangladesh} & 2023 & -- & 90.31 & 138 & -- & -- & Bangladesh Dataset \\ 
\midrule

EMDFNet \cite{li2024emdfnet} & 2024 & 97 & -- & 29.7 & 16.7 & RTX 3090 & GTSDB\\

\midrule
YOLO-PPA \cite{zhang2024yolo} & 2024 & 93.2 & -- & 2.7 & 145.4 & RTX 3090 & GTSDB \\ \midrule

\multirow{2}{*}{Enhanced YOLOv8 \cite{ibrahim2025enhancing}} & \multirow{2}{*}{2025} & \multirow{2}{*}{91.5} & \multirow{2}{*}{--} & \multirow{2}{*}{--} & 45 & RTX 3080 & \multirow{2}{*}{GTSRB} \\
 & &  &  &  & 10.4 & Jetson Nano &  \\ \midrule 

\multirow{3}{*}{\textbf{Ours proposed Framework}} & \multirow{3}{*}{2025} & \multirow{3}{*}{97.4} & 93.1 & \multirow{3}{*}{13.9} & 18 & Orin Nano & Local \cite{mexwell_pakistanitrafficsign_2025} \\

 & & &95.7& & 89 & Tesla P100 & Bangladesh dataset \cite{Tusher2020TrafficSignBangladesh} \\

 & & &88.4& & 110 & RTX 3090 & GTSRB \cite{Houben-IJCNN-2013} \\

\bottomrule
\end{tabular}%
}
\end{table}
\section{Lane Detection \& Departure Warning}

Lane detection is a fundamental component of ADAS, which allows vehicles to recognize road boundaries and maintain proper lane discipline.. Accurate lane detection not only supports autonomous path planning and lane keeping, but also plays a critical role in identifying unintended lane departures. A major cause of collisions is when drivers are fatigued, distracted, or driving in low visibility conditions. Building on this capability, a lane departure warning system continuously monitors the vehicle's position relative to the lane boundaries and issues timely alerts whenever the vehicle begins to drift unintentionally. 

\subsection{Lane Detection and Segmentation}
\begin{figure}
    \centering
    \includegraphics[width=0.8\linewidth]{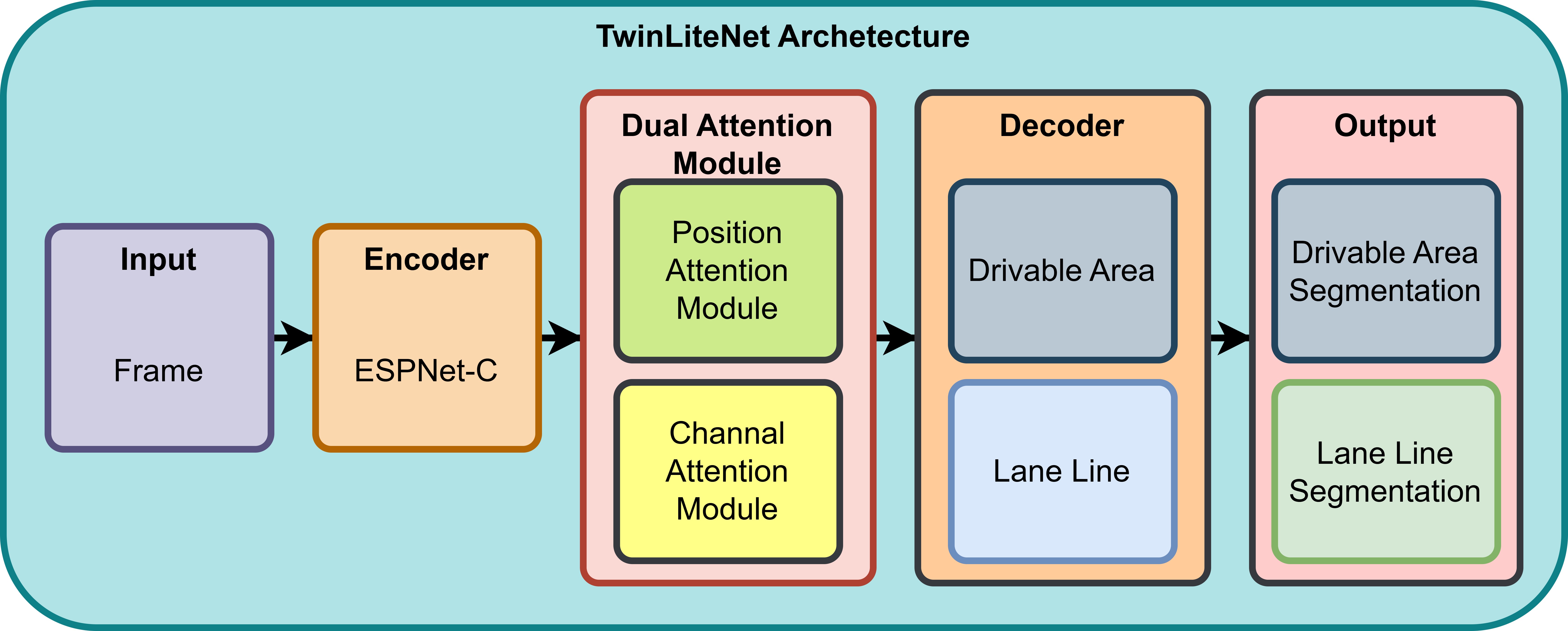}
    \caption{Block Diagram of Architecture of TwinLiteNet}
    \label{fig:modified}
\end{figure}

\begin{figure}
    \centering
    \includegraphics[width=0.8\linewidth]{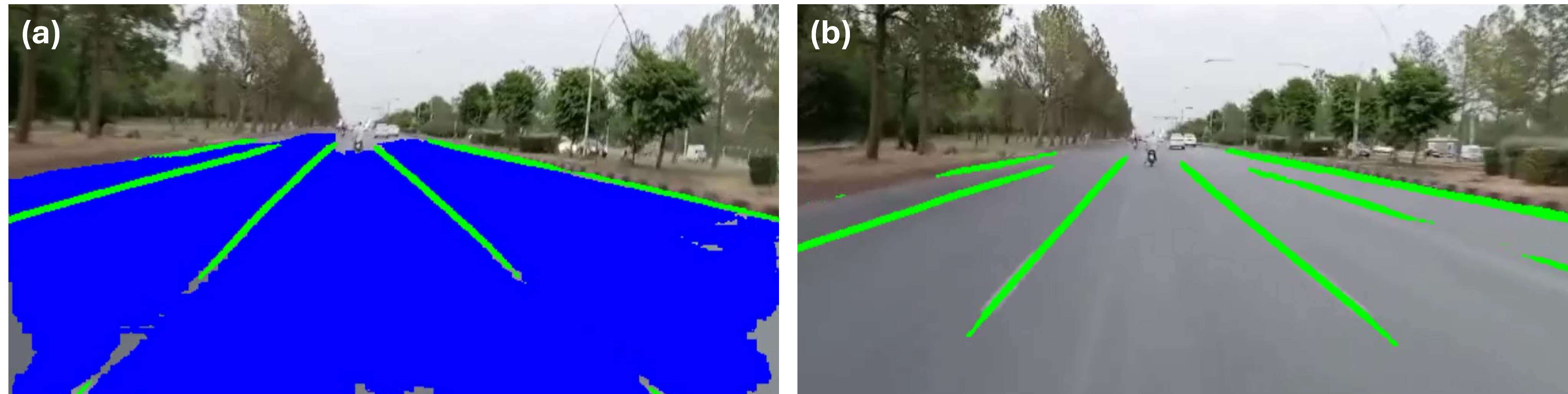}
    \caption{(a) Shows the orignal result of TwinLiteNet Model (b) shows the Lane Line Segmentation from TwinLiteNet Model}
    \label{fig:lane line}
\end{figure}

Block Diagram of Archetecture of the TwinLiteNet as shown in Figure~\ref{fig:modified}), was chosen as the main model for lane detection and segmentation in this work due to its lightweight design (0.44 million parameters). Figure~\ref{fig:lane line}(a) displays the TwinLiteNet results, with the drivable region segmented in blue and the lanes segmented in green. In this system, lane line segmentation is the main emphasis for lane departure analysis. The drivable area decoder was eliminated from the design with the remaining lane line segmentation head in order to optimize the model. This modification reduced the model size by approximately 3,000 parameters. The result of modified model is shown in Figure~\ref{fig:lane line}(b).
However, in real-world scenarios where lane markings are missing, obstructed, or faded, the system incorporates a fallback approach. In such cases, the segmented drivable area is used to approximate lane boundaries, which suggest that lane awareness can be maintained even when clear lane markings are not visible.

\begin{algorithm}
\caption{Custom Designed Lane Departure Warning (LDW) Algorithm}
\label{alg:LDW}
\KwIn{Input frame $I$}
\KwOut{Lane–departure warning signal and visualized output}

1: $I_r \gets \text{Resize}(I,640 \times 360)$ \

2: $I_{\text{low}} \gets \text{LowerHalf}(I_r, 640 \times 180)$ \

3: $I_M \gets$ TwinLiteNet($I_{\text{low}}$) \tcp*{Binary lane mask}
4: $I_C \gets \text{Canny}(I_M)$ \tcp*{Canny edges}
5: $I_H \gets \text{Hough}(I_C)$ \tcp*{Hough transform lines}
6: \text{Classify} $I_H$ \text{into } $\{H_L, H_R\}$ \text{based on slope} \

7: Define bounding box $B \in \mathbb{R}^{100 \times 160}$ at bottom-center of $I_{\text{low}}$ \

8: $P_{\text{veh}} \gets \text{center}(B)$ \

9: \textbf{if} $H_L \cap B \neq \varnothing$ \textbf{or} $H_R \cap B \neq \varnothing$ \

\quad \text{Trigger warning} \

10: \textbf{else} \

\quad  \text{No warning} \

11: Overlay $\{I_M, I_H, P_{\text{veh}}\}$ \text{on } $I_r$ with lane centers (left, right, $P_{\text{veh}}$)

\end{algorithm}

\subsection{Lane Departure Warning}

\begin{figure}[!t]
    \centering
    \includegraphics[width=0.8\linewidth]{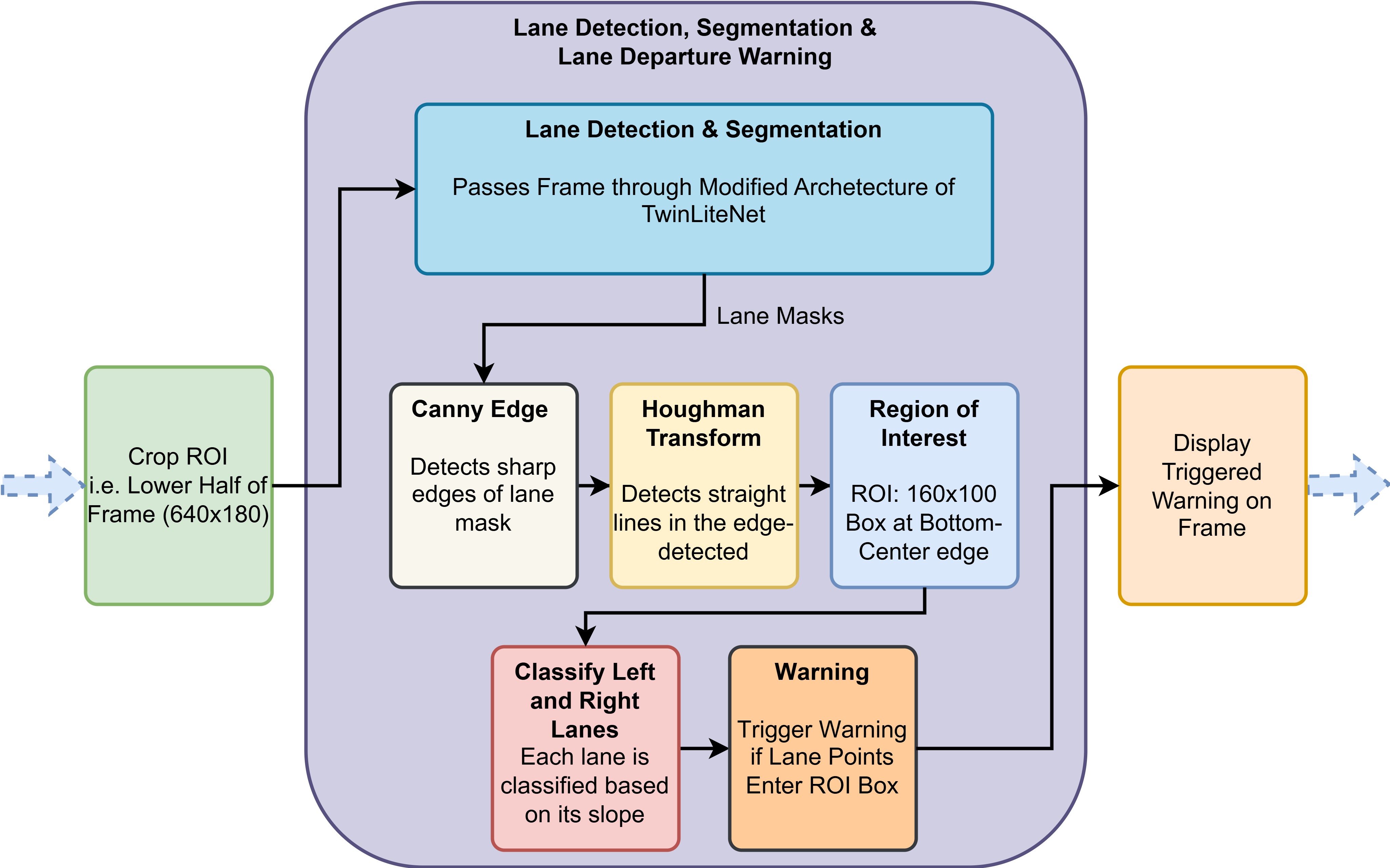}
    \caption{Lane Departure Warning Algorithm}
    \label{fig:LDW Algo}
\end{figure}

\begin{figure}
    \centering
    \includegraphics[width=0.5\linewidth]{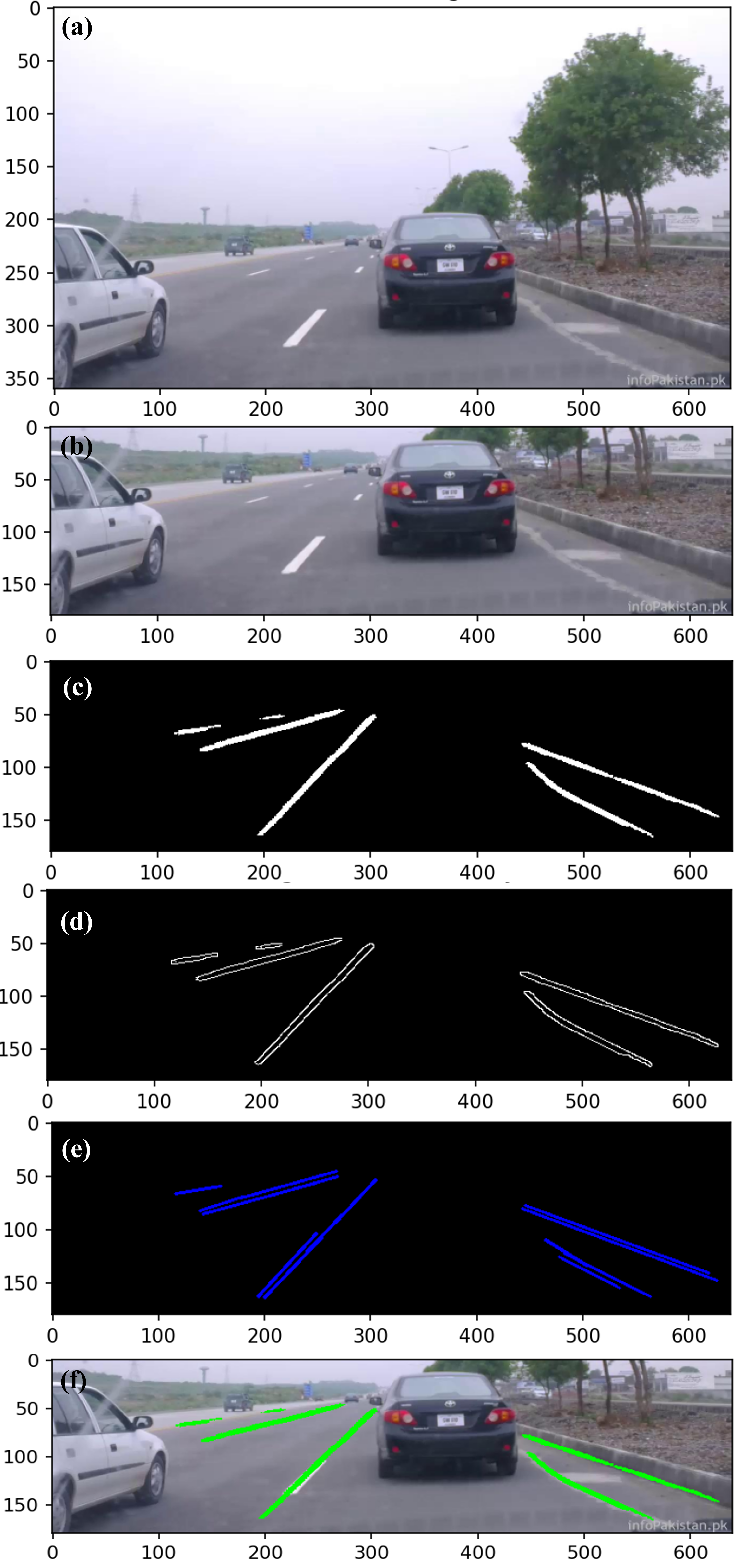}
    \caption{(a) Resizing Frame to 640x360 (b) Cropping Frame for getting ROI i.e. 640x180. (c) Lane Mask from Model. (d) Canny Edge Detection Output. (e) HoughLine Transform Output (f) Placing Mask on ROI}
    \label{fig:IPT}
\end{figure}

\begin{figure} 
    \centering
    \includegraphics[width=0.6\linewidth]{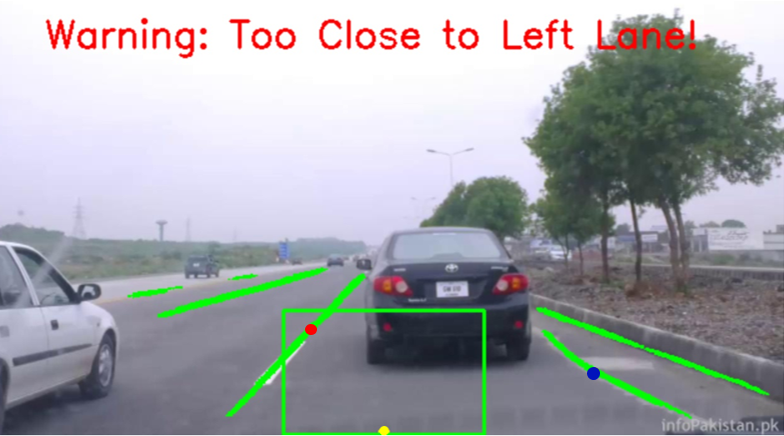}
    \caption{Lane Departure Warning. Red and Blue points indicating center of Right and Left lane respectively while yellow point indicating Vehicle position which is the bottom middle edge of frame. Left lane center comes in threshold window so the Warning Generated for Left lane.}
    \label{fig:final LDW}
\end{figure}

\begin{table}[]
\centering
\caption{Detailed Performance Metrics for Warning Detection Across Different Environmental Conditions}
\label{tab:combined_performance}
\small
\resizebox{\linewidth}{!}{%
\begin{tabular}{@{} l l c c c c c c @{}}
\toprule
\textbf{Scenario} & \textbf{Class} & \textbf{Precision} & \textbf{Recall} & \textbf{F-1} & \textbf{Pred. Frames} & \textbf{GT} & \textbf{Accuracy (\%)} \\ 
\midrule
\multirow{3}{*}{Day Time} 
    & No Warning    & 0.83 & 1.00 & 0.91 & 194 & 161 & \multirow{3}{*}{91.56} \\
    & Right Warning & 0.85 & 0.24 & 0.37 & 13  & 46  & \\
    & Left Warning  & 0.96 & 0.96 & 0.96 & 386 & 386 & \\
\cmidrule(lr){2-8}
\multicolumn{2}{r}{\textit{Scenario Totals:}} & \multicolumn{3}{l}{Total Frames: 593} & \multicolumn{2}{l}{Correct: 543} & \\
\midrule
\multirow{3}{*}{Night Time} 
    & No Warning    & 0.76 & 0.97 & 0.85 & 1850 & 1441 & \multirow{3}{*}{85.86} \\
    & Right Warning & 0.97 & 0.82 & 0.89 & 1189 & 1414 & \\
    & Left Warning  & 0.98 & 0.69 & 0.81 & 442  & 626  & \\
\cmidrule(lr){2-8}
\multicolumn{2}{r}{\textit{Scenario Totals:}} & \multicolumn{3}{l}{Total Frames: 3481} & \multicolumn{2}{l}{Correct: 2989} & \\
\midrule
\multirow{3}{*}{Rain} 
    & No Warning    & 1.00 & 0.99 & 0.99 & 1194 & 1209 & \multirow{3}{*}{97.09} \\
    & Right Warning & 0.44 & 0.93 & 0.60 & 63   & 30   & \\
    & Left Warning  & 1.00 & 0.49 & 0.65 & 17   & 35   & \\
\cmidrule(lr){2-8}
\multicolumn{2}{r}{\textit{Scenario Totals:}} & \multicolumn{3}{l}{Total Frames: 1274} & \multicolumn{2}{l}{Correct: 1237} & \\
\bottomrule
\end{tabular}%
}
\end{table}
\begin{figure}[!t]
    \centering
    \includegraphics[width=1\linewidth]{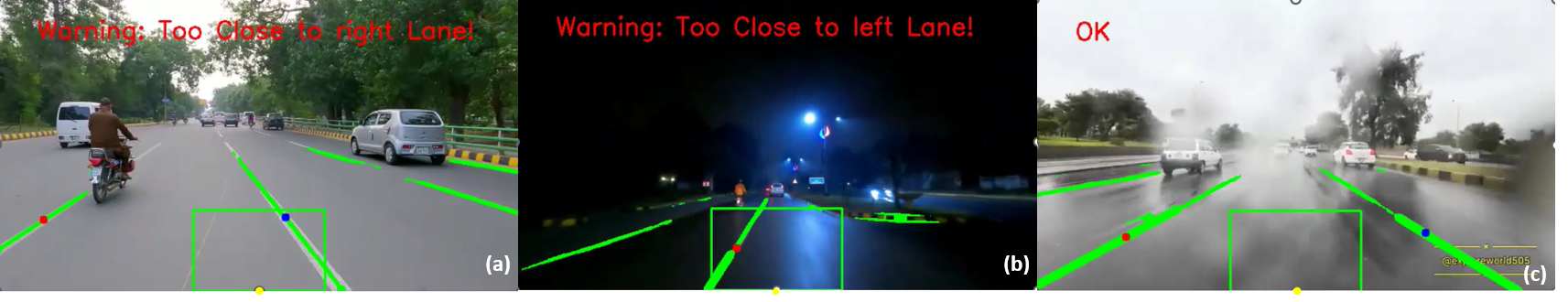}
    \caption{Evaluation of Lane Departure Algorithm in Different Scenarios i.e. (a) Day (b) Night and (c) Rain }
    \label{fig:LDW Evaluation}
\end{figure}

The lane departure warning (LDW) system is based on the algorithm shown in Figure~\ref{fig:LDW Algo}. The output of the modified TwinLiteNet model is used as the foundation for this system. The propoased method segments the lanes on the lower half of each input frame. As outlined in the stepwise algorithmic workflow (Algorithm~\ref{alg:LDW}), the segmented lane mask undergoes a sequence of computer vision operations (discussed in the section followed by) for reliable lane departure detection.

Firstly, the original frame is captured and resized to $640 \times 360$ pixels, as shown in Figure~\ref{fig:IPT}(a). Next, the lower half of this frame is cropped and further reduced to $640 \times 180$ pixels (Figure~\ref{fig:IPT}(b)). This cropped region is then processed by the updated TwinLiteNet model, which generates a binary lane mask in real time highlighting the lane markings clearly, as shown in Figure~\ref{fig:IPT}(c). This binary mask serves as the input to the lane departure detection algorithm. Sharp lane edges are first extracted using Canny edge detection (Figure~\ref{fig:IPT}(d)), and straight lines are then detected using the Hough Transform (Figure~\ref{fig:IPT}(e)). Based on their slope, the detected lines are classified as either left or right lane boundaries. A $100 \times 160$ pixel box is defined at the bottom center of the frame to approximate the vehicle's position. The algorithm flags a potential lane departure whenever any of the classified left or right lane lines intersect this box.

The visual warning is superimposed on the output frame as shown in Figure~\ref{fig:final LDW}. The center of the left and right lanes are indicated by the red and blue points, while the yellow point shows the vehicle's position. In this example (Figure~\ref{fig:final LDW}), left-lane departure warning is triggered when the center of the left lane falls within the threshold window. The proposed hybrid approach combines deep learning and classical computer vision techniques for edge and line analysis to achieve a processing speed of 32 frames per second on the Jetson Orin Nano. The system is practical for embedded edge deployment in resource-constrained environments.

The reliability of the system was evaluated against a manually curated ground truth dataset, where each frame was annotated as having no warning, a left warning, or a right warning. Consistent performance across a variety of environmental conditions can be seen by comparing the system's predictions to this ground truth, as detailed in Table~\ref{tab:combined_performance}. During daytime driving (Figure~\ref{fig:LDW Evaluation}(a)), the system achieves an accuracy of 91.56\% under normal illumination, while maintaining 85.86\% accuracy in nighttime scenarios (Figure~\ref{fig:LDW Evaluation}(b)) despite the presence of headlight glare and reduced visibility. Notably, the system reached its highest accuracy of 97.09\% during rainy conditions (Figure~\ref{fig:LDW Evaluation}(c)), where it successfully suppressed false alerts during stable straight-line driving. These findings highlight the system's robustness and its ability to provide dependable lane perception even under adverse weather conditions.

\subsection{Fallback Mechanism for Lane Departure Warning}
\begin{figure}[]
    \centering
    \includegraphics[width=1\linewidth]{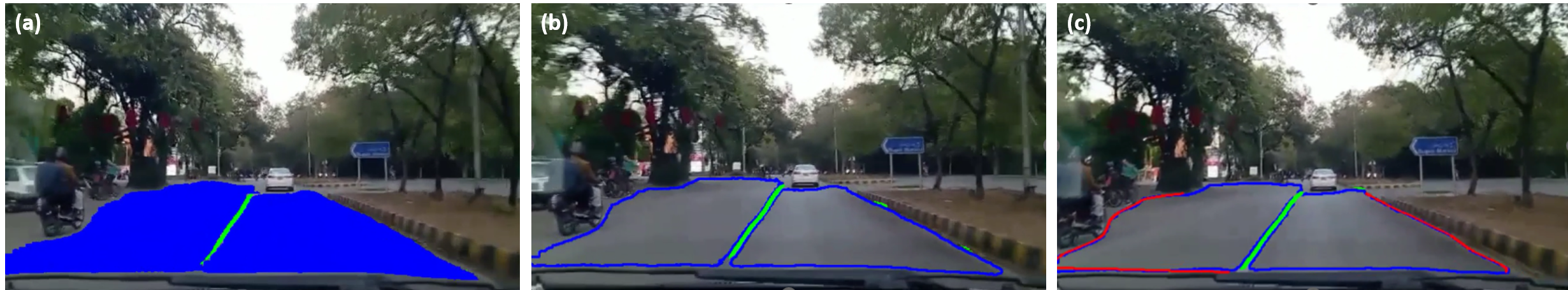}
    \caption{(a) Drivable Area Segmentation (b) Edges of Drivable Area Segmentation (c) Classify Left and Right Edge of Drivable Area}
    \label{fig:DAS1}
\end{figure}

\begin{figure}
    \centering
    \includegraphics[width=1\linewidth]{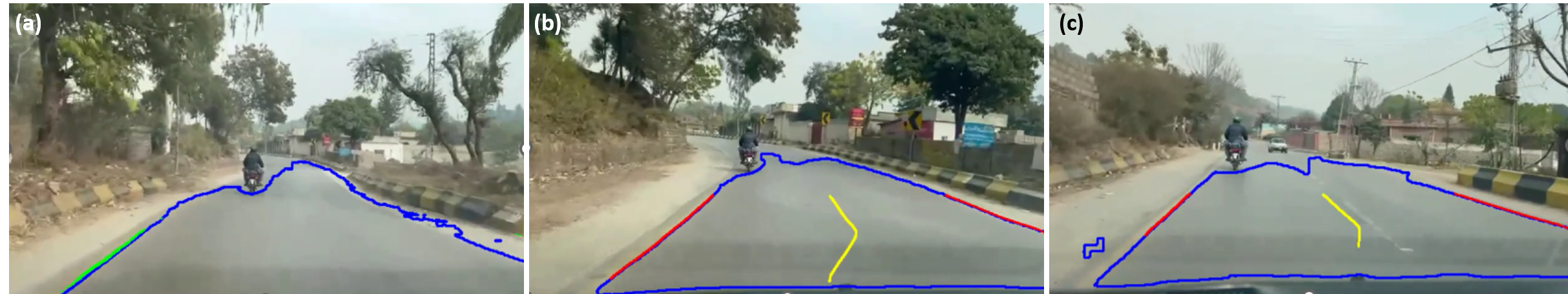}
    \caption{(a) shows the Edges of Drivable Area Segmentation (b) and (c) shows the Lane Approximation using midpoint of Left and Right Edge of Drivable Area}
    \label{fig:DAS2}
\end{figure}

\begin{figure} 
    \centering
    \includegraphics[width=1\linewidth]{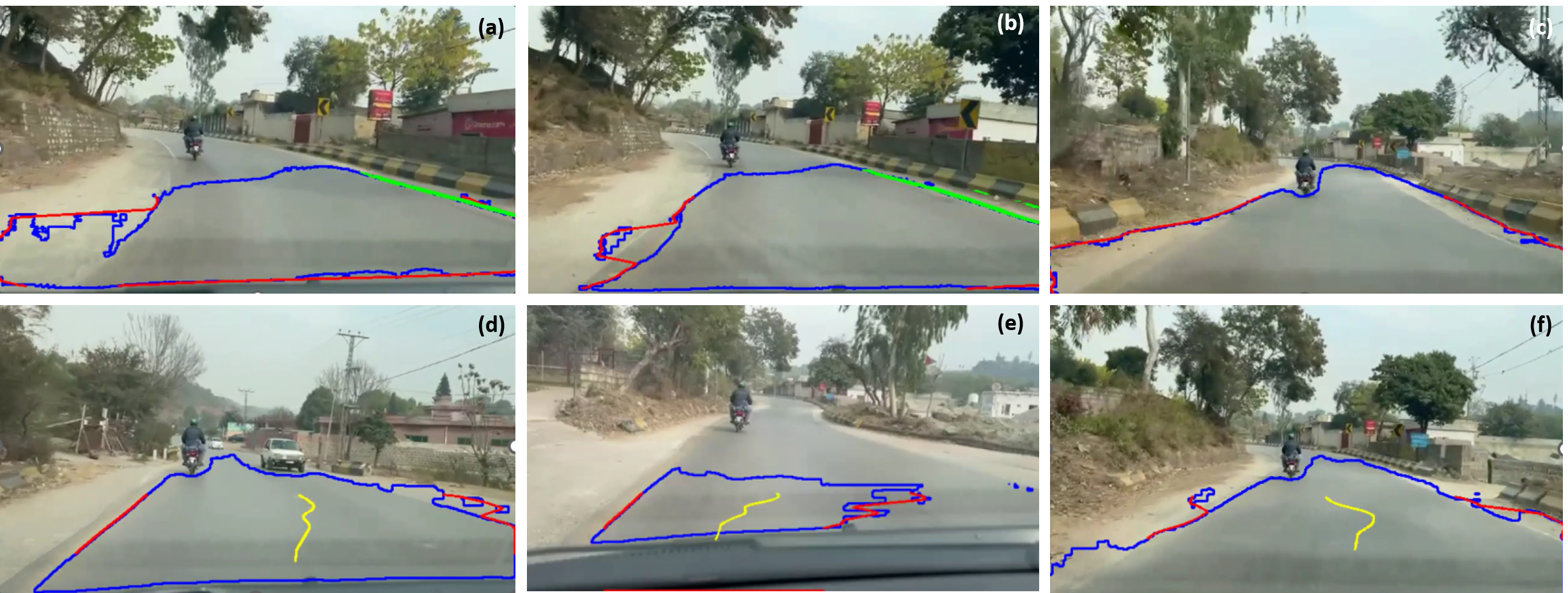}
    \caption{Noise in Drivable Area Segmentation due to which lane approximation becomes unstable}
    \label{fig:DAS3}
\end{figure}

\begin{figure}[!]
    \centering
    \includegraphics[width=1\linewidth]{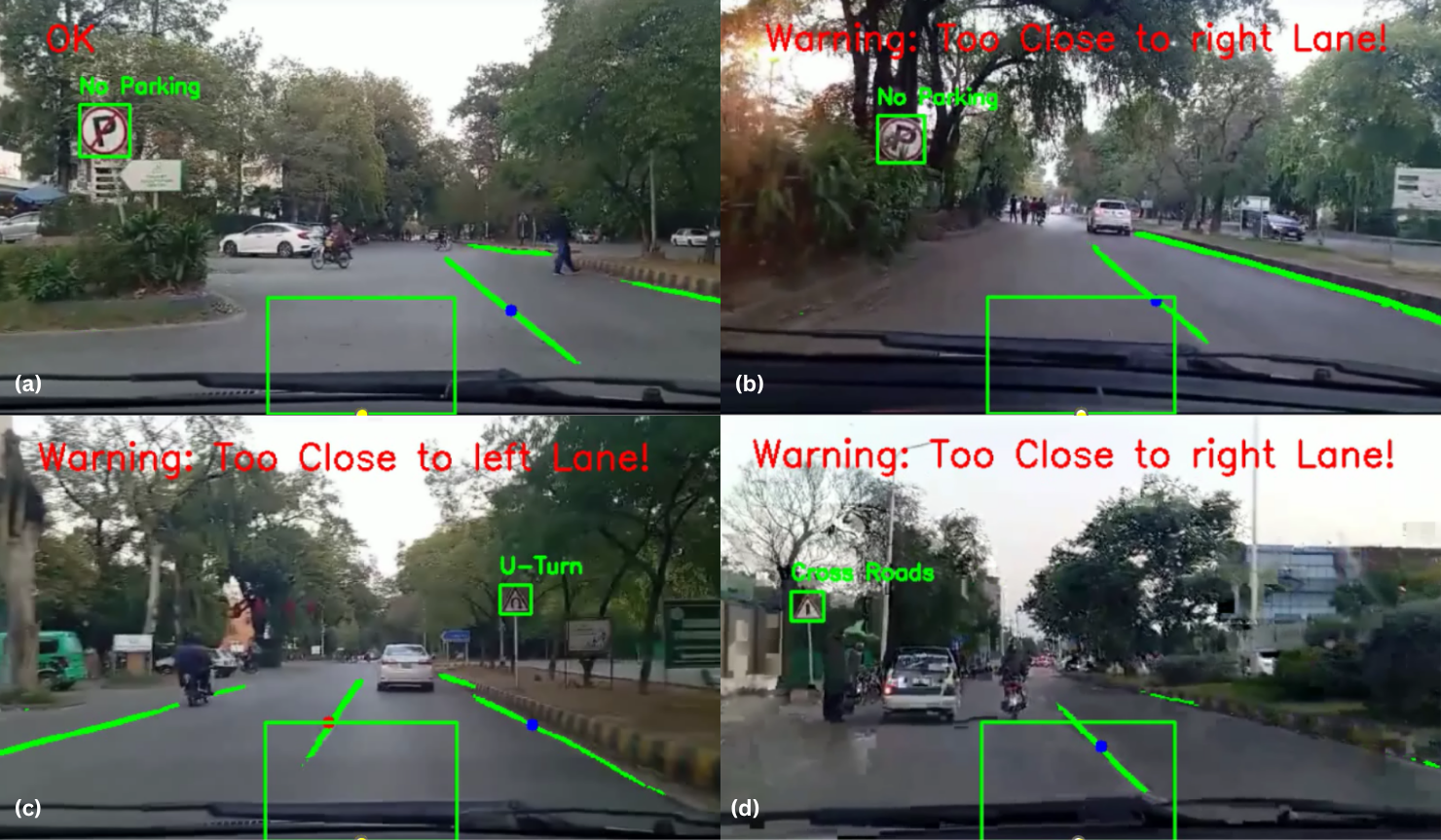}
    \caption{Lane departure warning with traffic sign detection and recognition. (a) normal driving with a "No Parking" sign detected and no lane departure. (b) detection of a "No Parking" sign and warning for being too close to the right lane. (c) a "U-Turn" sign and warning for closeness to the left lane. (d) detection of a "Cross Roads" sign and also indicates a lane departure warning for drifting too close to the right lane}
    \label{fig:LDW&TSD}
\end{figure}

\begin{table}[]
\centering
\caption{Comparison of different models for lane and traffic sign analysis. 
Here, \textbf{\checkmark } indicates that the model provides the corresponding functionality, 
whereas \textbf{X} indicates that the functionality is not supported.}
\label{tab:comparison_matrix_2}
\resizebox{\linewidth}{!}{%
\begin{tabular}{l c c c c c c c c c}
\toprule
\textbf{Model} & \textbf{Year} & \textbf{TSD} & \textbf{TSR} & \textbf{Object Det} & \textbf{LD} & \textbf{DAS} & \textbf{LDW} & \textbf{Edge Device} & \textbf{Real-Time (FPS)} \\
\midrule

\multicolumn{10}{l}{\textbf{Multi-task Object Detection \& Lane-Aware Models}}\\
\midrule
YOLOP \cite{wu2022yolop} & 2021 & \XSolid & \XSolid & \checkmark & \checkmark & \checkmark & \XSolid & Jetson TX2 & 23 \\
HybridNet \cite{vu2022hybridnets}& 2022 & \XSolid & \XSolid & \checkmark & \checkmark & \checkmark & \XSolid & \XSolid & \XSolid \\
YOLOPv2 \cite{han2022yolopv2} & 2022 & \XSolid & \XSolid & \checkmark & \checkmark & \checkmark & \XSolid & \XSolid & \XSolid \\
CENTERNETS \cite{chen2023centerpnets} & 2023 & \XSolid & \XSolid & \checkmark & \checkmark & \checkmark & \XSolid & \XSolid & \XSolid \\
YOLOPv3 \cite{zhan2024multi}  & 2024 & \XSolid & \XSolid & \checkmark & \checkmark & \checkmark & \XSolid & \XSolid & \XSolid \\
YOLOPX \cite{batista2023application} & 2024 & \XSolid & \XSolid & \checkmark & \checkmark & \checkmark & \XSolid & \XSolid & \XSolid \\
YOLOP-DN \cite{tan2024multi} & 2024 & \XSolid & \XSolid & \checkmark & \checkmark & \checkmark & \XSolid & \XSolid & \XSolid \\
YOLOM \cite{wang2024you} & 2024 & \XSolid & \XSolid & \checkmark & \checkmark & \checkmark & \XSolid & \XSolid & \XSolid \\

\midrule
\multicolumn{10}{l}{\textbf{Lane Focused Models}}\\
\midrule
LLDNET \cite{khan2022lldnet} & 2022 & \XSolid & \XSolid & \XSolid & \checkmark & \XSolid & \XSolid & \XSolid & \XSolid \\
LATR \cite{luo2023latr} & 2023 & \XSolid & \XSolid & \XSolid & \checkmark & \XSolid & \XSolid & \XSolid & \XSolid \\
TwinLiteNet \cite{che2023twinlitenet} & 2023 & \XSolid & \XSolid & \XSolid & \checkmark & \checkmark & \XSolid & Jetson Xavier NX & 60 \\
SRLANE \cite{chen2024sketch} & 2024 & \XSolid & \XSolid & \XSolid & \checkmark & \XSolid & \XSolid & \XSolid & \XSolid \\
CURVEFORMER++ \cite{bai2025curveformer++} & 2025 & \XSolid & \XSolid & \XSolid & \checkmark & \XSolid & \XSolid & \XSolid & \XSolid \\

\midrule
\multicolumn{10}{l}{\textbf{Traffic-Sign Detection \& Recognition Models}}\\
\midrule
YOLOv5 \cite{dang2023improved} & 2023 & \checkmark & \checkmark & \XSolid & \XSolid & \XSolid & \XSolid & \XSolid & \XSolid \\
EMDFNet \cite{li2024emdfnet} & 2024 & \checkmark & \checkmark & \XSolid & \XSolid & \XSolid & \XSolid & \XSolid & \XSolid \\
YOLO-PPA \cite{zhang2024yolo} & 2024 & \checkmark & \checkmark & \XSolid & \XSolid & \XSolid & \XSolid & \XSolid & \XSolid \\
YOLO-TS \cite{chen2024yolo} & 2024 & \checkmark & \checkmark & \XSolid & \XSolid & \XSolid & \XSolid & Jetson Orin AGX & 38.8\\
Enhanced YOLOv8 \cite{ibrahim2025enhancing} & 2025 & \checkmark & \checkmark & \XSolid & \XSolid & \XSolid & \XSolid & Jetson Nano & 10.4\ \\
MH-YOLO \cite{zhang2025mh} & 2025 & \checkmark & \checkmark & \XSolid & \XSolid & \XSolid & \XSolid & \XSolid & \XSolid \\
YOLO-BS \cite{zhang2025yolo} & 2025 & \checkmark & \checkmark & \XSolid & \XSolid & \XSolid & \XSolid & \XSolid & \XSolid \\

\midrule
\multicolumn{10}{l}{\textbf{Integrated Lane \& Traffic-Sign Models}}\\
\midrule
Improved YOLOv8 \cite{elankumaranimproved}  & 2023 & \checkmark & \checkmark & \XSolid & \checkmark & \XSolid & \XSolid & \XSolid & \XSolid \\
Lane and Traffic Sign Detection (YOLOv8) \cite{yaamini2025lane} & 2025 & \checkmark & \checkmark & \XSolid & \checkmark & \XSolid & \XSolid & \XSolid & \XSolid \\
Advancing Autonomous Vehicle Intelligence \cite{sah2025advancing} & 2025 & \checkmark & \checkmark & \XSolid & \checkmark & \XSolid & \XSolid & \XSolid & \XSolid \\

\midrule
\multicolumn{10}{l}{\textbf{Proposed Multi-Task Model (This Work)}}\\
\midrule
Ours (TSD\&R + Custom Designed LDW Algo) & 2025 & \checkmark & \checkmark & \XSolid & \checkmark & \XSolid & \checkmark & Jetson Orin Nano & 15 \\

\bottomrule
\end{tabular}%
}
\end{table}

Lane line segmentation occasionally fails in real-world driving circumstances because of occlusions, dim illumination, damaged roads, or faded markers. TwinLiteNet's secondary output, which offers drivable region segmentation, is suggested as a fallback strategy to handle such situations.

As seen in Figure~\ref{fig:DAS1}(a), this drivable area segmentation spans the whole road surface and is nonetheless quite reliable even in the absence of lane markers. Figure~\ref{fig:DAS1}(b) and Figure~\ref{fig:DAS2}(a) illustrate how the outside limits of the driveable area are retrieved in the fallback mode. As seen in Figure~\ref{fig:DAS1}(c), these boundaries are then categorized as left and right road edges according to their respective positions inside the frame. After classification, midpoints are determined across the frame's vertical axis between the appropriate left and right border points. As seen in Figure~\ref{fig:DAS2}(b) and Figure~\ref{fig:DAS2}(c), connecting these midpoints creates an estimated centerline of the driveable route that may be used to estimate the location of the vehicle and identify possible lane departure.

\begin{figure}[]
    \centering
    \includegraphics[width=0.7\linewidth]{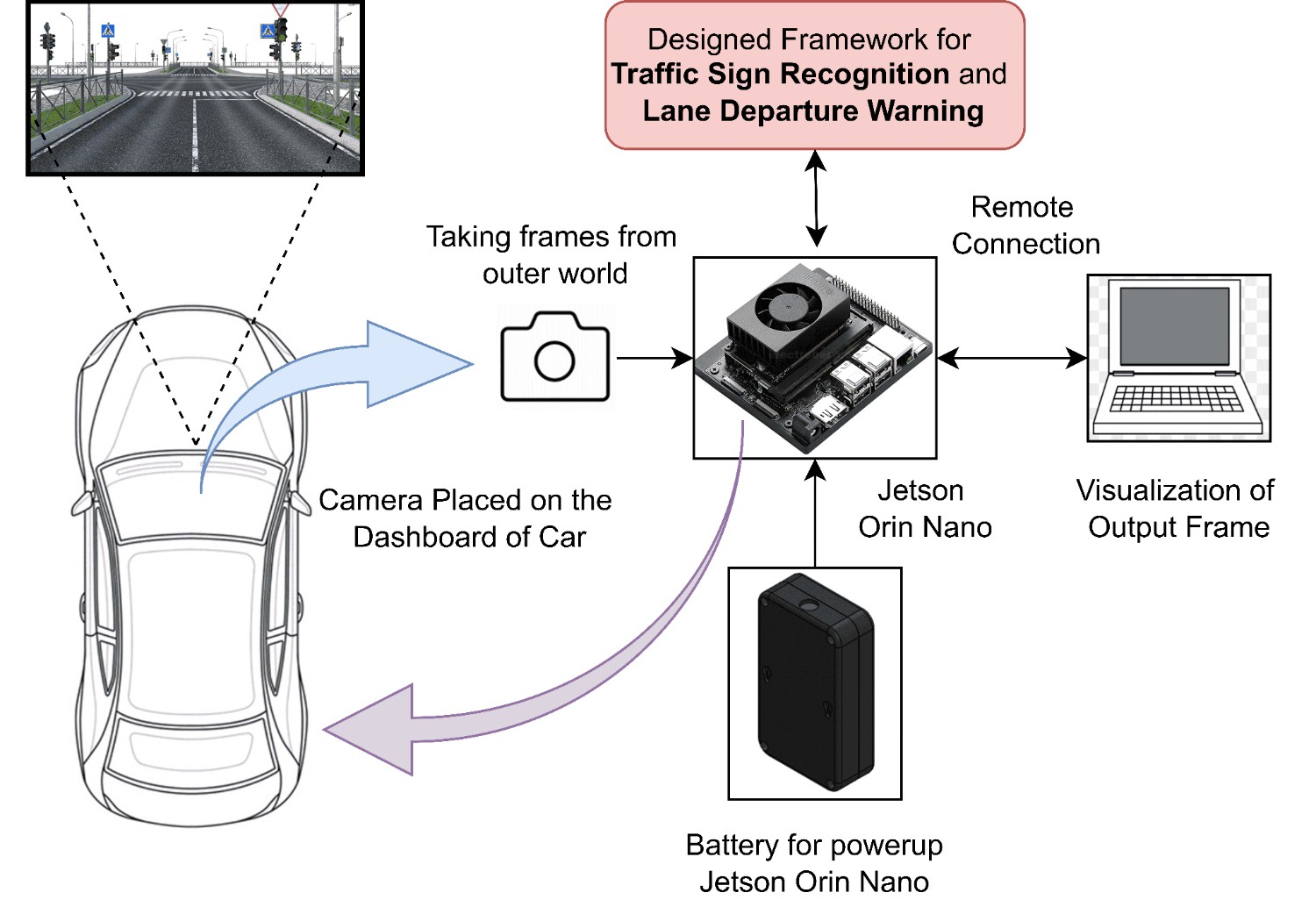}
	\captionsetup{justification=justified,singlelinecheck=false}
    \caption{ Overview of proposed experimental setup and data flow. A camera mounted on the vehical dashboard capture real time road scenes. the captured frames are processed on jetson orin nano to perform traffic sign recognition and lane departure warning. the processd result are trasmitted via remote connection for visualization, while warning information delivered back to the driver. The Jetson Orin Nano is powered using external battery.}
    \label{fig:experimental_setup schematics}
\end{figure}

Initial tests using the pre-trained model show that this method provides a reasonable approximation in many scenarios. However, the quality of the estimated centerline is directly dependent on the accuracy of the drivable area segmentation. In cases where the segmentation is inaccurate, especially around curves or road edges, the resulting midpoints and lane estimations tend to be unstable as shown in Figure~\ref{fig:DAS3}. Despite this, the approach demonstrates a viable proof of concept. With proper model fine-tuning and domain-specific training, this fallback strategy has the potential to serve as a reliable alternative in lane-degraded environments.

In many situations, this approach yields a fair approximation according to preliminary tests conducted with the pre-trained model. However, the precision of the drivable area segmentation directly affects the quality of the calculated centerline. As shown in Figure~\ref{fig:DAS3}, the resultant midpoints and lane estimates are frequently unstable when the segmentation is incorrect usually near bends or road borders. In lane-degraded contexts, this fallback technique may be a dependable substitute with the right model fine-tuning and domain-specific training.

\section{Integration of Lane Departure Warning and Traffic Sign Detection \& Recognition}

The proposed system integrates lane departure warning (LDW) and traffic sign detection and recognition (TSD/TSR) into a single real-time Advanced Driver Assistance System (ADAS). The live camera frames are divided into upper and lower halves. The upper half is processed using YOLOv8 for traffic sign detection and MobileNetV3-small for accurate recognition, while the lower half is processed by the modified version of TwinLiteNet model for lane segmentation. A custom lane departure algorithm calculates the vehicle's position relative to lane boundaries using the bottom-center pixel as a reference. When the vehicle drifts too close to lane edges, clear visual warnings are issued to alert the driver immediately. Figure~\ref{fig:LDW&TSD} shows representative outputs for different driving scenarios.

To evaluate the novelty and effectiveness of the proposed system, Table~\ref{tab:comparison_matrix_2} compares existing models for lane detection, traffic sign detection, and their combination. Most prior works specialize in a single task or lack support for real-time processing and edge deployment. For example, YOLOv5, EMDFNet, and YOLO-PPA achieve high performance for traffic sign recognition but do not provide lane departure warnings. On the other hand, lane detection models such as LLDNet, SRLANE, and TwinLiteNet focus on lane segmentation but do not detect traffic signs, thus they have limited situational awareness. From Table~\ref{tab:comparison_matrix_2}, it is evident that our system is one of the few solutions that simultaneously perform both lane departure warning and traffic sign recognition while being deployable on edge devices. Moreover, our framework achieves 15 FPS on the Jetson Orin Nano, aligning with prior studies that consider similar frame rates sufficient for real-time embedded TSR applications \cite{elmanaa2023efficient}, \cite{santos2020real},\cite{chenreal}, \cite{mccarthy2019vision}. The combination of accuracy, edge compatibility, and dual task integration makes our approach more practical and robust than existing models.

\section{Real Time Implementation of Framework on Edge Devices}

To validate the real-time performance of the ADAS framework on the edge device, a testing setup was established. Schematic diagram of our experimental setup shown in Figure.~\ref{fig:experimental_setup schematics}. Photos of the Setup are shown in Figure S1 of Supplementary Information. It consist of three major components: the edge computing device (NVIDIA Jetson Orin Nano), an external camera for real-time visual input, and monitor for the processed output (a local web server was define to see the visual output). The entire setup was powered through a custom built portable battery pack (Volatge rating: 12 V, Current Rating: 2.5 A). Photos of real-time driving experiments, illustrating traffic sign detection and lane detection, are presented in Figures S2 and S3 of the Supplementary Information.

An Intel RealSense D435 RGB-D camera was used for video capture, with the factory-calibrated intrinsic and extrinsic parameters provided by the manufacturer. The major deployment target was Jetson Orin Nano which is a very efficient and powerful compact edge AI device. It delivered high performance as compared to the other tested platforms (Results shown in Table S2 of Supplementry Information). It achieved an interference speed of 18 FPS for the traffic sign recognition module and 32 FPS for the lane departure module with ONNX optimization. The system achieved a stable 15 FPS end-to-end processing rate under different lighting conditions. 

Overall, Real-time performance is enabled by the fusion of lightweight deep learning models, efficient data flow, and hardware acceleration through ONNX and TensorRT. This real-time implementation establishes a foundation for future enhancements toward fully autonomous driving systems, improving situational awareness, safety, and real-world deployability.

The proposed framework showed reliable real-time performance, however, there are can be improvements such as extending the ADAS functionality to include speed-limit compliance, driver state monitoring, and fallback mechanisms for lane-center
estimation in complex road geometries. Dataset expansion and domain adaptation techniques can reduce retraining requirements across diverse environments.

\section{Conclusion}
This work demonstrated a unified, edge-deployable ADAS that performs lane-departure warning and traffic-sign detection/recognition in real time on automotive-class hardware. By integrating both functions within a lightweight framework and validating end-to-end on a Jetson Orin Nano, we showed that practical throughput can be achieved without sacrificing accuracy, including high mAP for detection and strong recognition accuracy on a region-specific datasets augmented from scarce originals. The resulting system addresses a common gap in the literature, where ADAS capabilities are developed as isolated modules or evaluated only on desktop-class GPUs, and provides a credible path toward low-cost deployment in settings with limited compute and non-standardized signage.
 
\section*{Acknowledgments}
The authors acknowledge support from the National University of Sciences and Technology (NUST) for conducting this research.
 
\bibliographystyle{unsrtnat}
\bibliography{bibliography}
 
\clearpage
\appendix
 
\setcounter{section}{0}
\setcounter{table}{0}
\setcounter{figure}{0}
\renewcommand{\thesection}{S\arabic{section}}
\renewcommand{\thetable}{S\arabic{table}}
\renewcommand{\thefigure}{S\arabic{figure}}
\renewcommand{\theHsection}{S\arabic{section}}
\renewcommand{\theHtable}{S\arabic{table}}
\renewcommand{\theHfigure}{S\arabic{figure}}
 
\begin{center}
    {\Large\bfseries Supplementary Information to:}\\[0.6em]
    {\large\bfseries Edge Deployable Lightweight Framework for Real-Time Driver Safety System with Integrated Lane Departure and Traffic Sign Recognition}
\end{center}
\vspace{1em}
 
\section*{Introduction}
 
This supplementary document accompanies the article ``Edge Deployable Lightweight Framework for Real-Time Driver Safety System with Integrated Lane Departure and Traffic Sign Recognition,'' in which we presented a unified, edge-deployable ADAS framework that simultaneously performs lane-departure warning (LDW) and traffic-sign detection and recognition (TSD/TSR), and validated the proposed system on automotive-class devices. While the original article provides an overview of the methodology and key results, several additional details are necessary for a complete understanding and replication of the experiments.
 
Specifically, this document provides extended descriptions of the experimental setup, parameter configurations, additional ablation studies, and supporting results that could not be included in the main manuscript due to space limitations. These materials aim to enhance transparency, reproducibility, and provide a deeper insight into the experimental procedures and results reported in the main paper.

\section{Dataset Classes}
\label{sec:s1}
 
Pakistani traffic signs dataset covering 35 classes listed in Table~\ref{tab:dataset}, were obtained from Kaggle \cite{mexwell_pakistanitrafficsign_2025}.
 
\begin{table}[!htbp]
\centering
\caption{List of 35 Traffic Sign Classes in the Pakistani Traffic Sign Dataset \cite{mexwell_pakistanitrafficsign_2025} (Kaggle)}
\label{tab:dataset}
\small
\begin{tabular}{clcl}
\toprule
\textbf{Class} & \textbf{Traffic Sign} & \textbf{Class} & \textbf{Traffic Sign} \\
\midrule
0  & Bridge Ahead           & 18 & Slow \\
1  & Cross Roads            & 19 & Speed Breaker Ahead \\
2  & Give Way               & 20 & Speed Limit (20 km/h) \\
3  & Left Bend              & 21 & Speed Limit (25 km/h) \\
4  & No Horns               & 22 & Speed Limit (30 km/h) \\
5  & No Mobile Allowed      & 23 & Speed Limit (40 km/h) \\
6  & No Overtaking          & 24 & Speed Limit (45 km/h) \\
7  & No Parking             & 25 & Speed Limit (50 km/h) \\
8  & No U-Turn              & 26 & Speed Limit (60 km/h) \\
9  & No Left Turn           & 27 & Speed Limit (65 km/h) \\
10 & No Right Turn          & 28 & Speed Limit (70 km/h) \\
11 & Parking                & 29 & Speed Limit (80 km/h) \\
12 & Pedestrians            & 30 & Steep Descent \\
13 & Railway Crossing       & 31 & Stop 1 \\
14 & Right Bend             & 32 & Stop 2 \\
15 & Road Divides           & 33 & U-Turn \\
16 & Roundabout Ahead       & 34 & Zigzag Road Ahead \\
17 & Sharp Right Turn       &    & \\
\bottomrule
\end{tabular}
\end{table}

\section{Testing and Experimental Setup}
\label{sec:s2}
 
To validate the real-time performance of the deployed ADAS framework on the Jetson Orin Nano, a comprehensive testing setup was established. The testing setup consisted of three major components: the edge computing device (Jetson Orin Nano), an external camera for real-time visual input, and a local web server for live monitoring of the processed output. The entire setup was powered through a portable battery pack to ensure mobility and field testing feasibility.
 
\begin{figure}[!ht]
    \centering
    \includegraphics[width=1\linewidth]{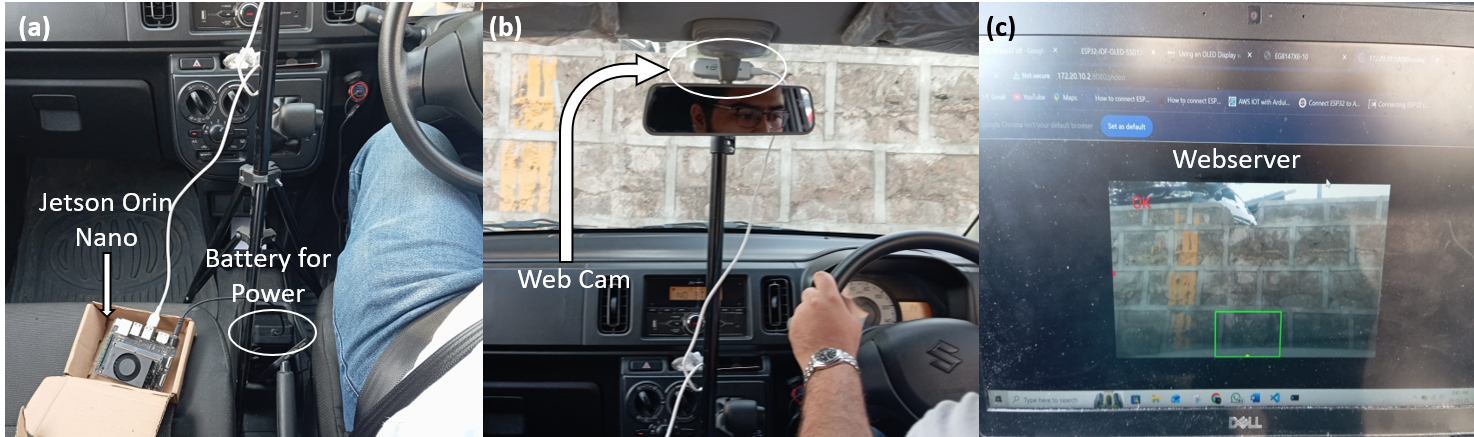}
    \caption{Experimental setup for real-time testing of the ADAS framework on Jetson Orin Nano. (a) Jetson Orin Nano powered through a portable battery. (b) Camera mounted on a tripod connected to the Orin Nano. (c) Local web server visualization of the real-time stream.}
    \label{fig:experimental_setup}
\end{figure}

As illustrated in Figure~\ref{fig:experimental_setup}, the Jetson Orin Nano was powered by a battery backup and connected via USB to an external camera mounted on a tripod stand for stable video acquisition. An Intel RealSense D435 RGB-D camera was used for video capture, employing the factory-calibrated intrinsic and extrinsic parameters provided by the manufacturer. The Orin Nano hosted the ADAS framework, performing both lane departure warning and traffic sign recognition tasks simultaneously. To visualize the processed results in real-time, a lightweight local web server was deployed on the device. The server streamed the inference output over the local network, enabling visualization on any connected device such as a laptop or mobile phone.
 
\begin{figure}[!ht]
    \centering
    \includegraphics[width=1\linewidth]{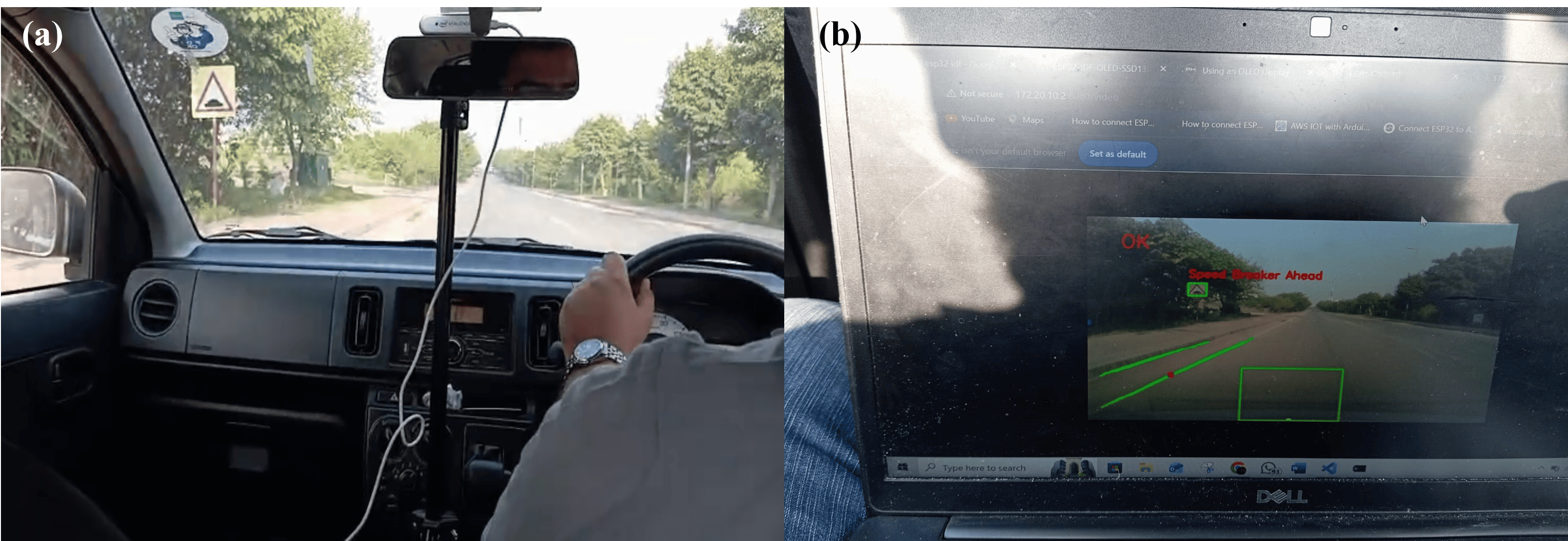}
    \caption{Qualitative real-time visualization of the proposed ADAS system. (a) displays the live camera feed from the test environment, and (b) illustrates the processed frame showing accurate lane detection and identification of the ``Speed Breaker Ahead'' traffic sign.}
    \label{fig:realtime_results}
\end{figure}

\begin{figure}[!h]
    \centering
    \includegraphics[width=1\linewidth]{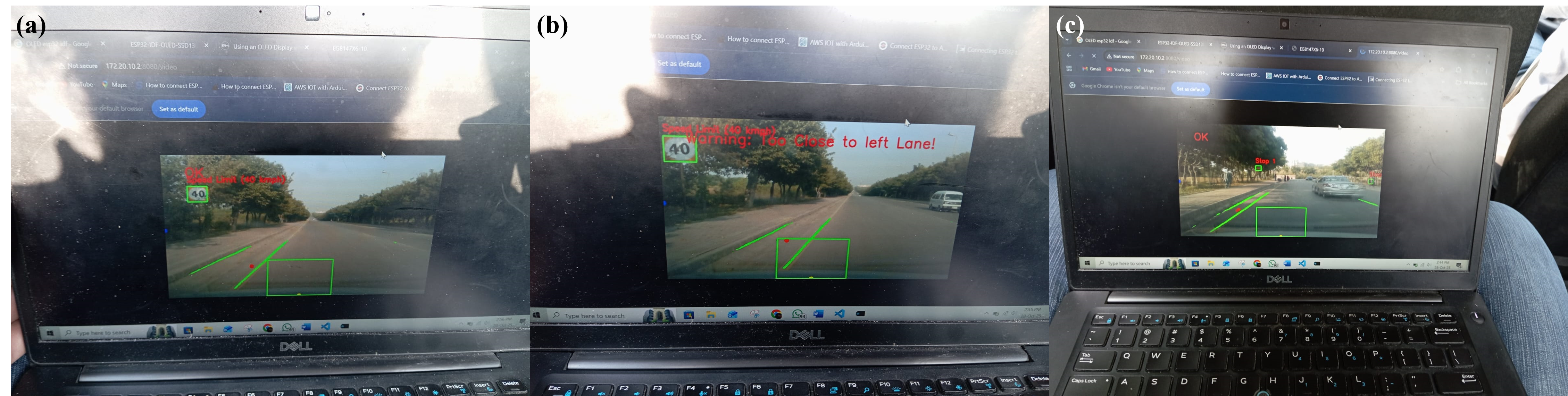}
    \caption{Qualitative real-time results of the proposed ADAS framework. (a) shows detection of a ``Speed Limit 40'' sign with no lane marking warning, indicating that the vehicle is at a safe distance. (b) illustrates the same detected sign when the vehicle moves closer to the lane boundary, triggering a lane departure warning on the frame. (c) presents detection of a ``Stop'' sign with no warning generated, as the vehicle remains within the safe zone.}
    \label{fig:realtime_results_2}
\end{figure}

During testing, the framework successfully processed live road video streams, generating lane departure and traffic sign recognition results in real time. The integrated system achieved a stable 15 FPS end-to-end processing rate, maintaining low latency and reliable detection even under varying lighting conditions. The qualitative results of real-time testing are shown in Figure~\ref{fig:realtime_results} and Figure~\ref{fig:realtime_results_2}, where detected lanes with warning and recognized traffic signs are superimposed on the live video feed.
 
Other than the Jetson Orin Nano, the lane departure module was also tested on other edge devices such as the Jetson Xavier, Jetson Nano, and Raspberry Pi. As per Table~\ref{tab:combined_results}, the Jetson Orin Nano delivered higher performance compared to the other tested platforms. It achieved an inference speed of 18 FPS for the traffic sign recognition module and 32 FPS for the lane departure module with ONNX optimization. The system maintained a real-time processing speed of 15 FPS when both modules were integrated into a single pipeline. This outcome confirms that the Jetson Orin Nano is well matched for embedded applications where low latency and real-time responsiveness are critical.
 
For comparison, the proposed system was also tested on other computing resources. First, it was tested on a general-purpose laptop (Dell Latitude, Core i5 8th Gen with 16 GB RAM and a 3.6 GHz processor, without GPU). It supported quantized models for running both modules, but its performance was comparatively lower than the other systems: the lane departure module ran at 9.5 FPS, the traffic sign recognition module ran at 7 FPS, and the combined pipeline operated at 5.5 FPS. Additionally, the lane module was tested on other NVIDIA Jetson platforms, namely the Jetson Nano and Jetson Xavier. On the Jetson Xavier, the lane departure module was tested with a PyTorch model, achieving 16 FPS. On the Jetson Nano (optimized with TensorRT), the same module achieved 11.1 FPS. These results highlight the importance of choosing appropriate hardware and leveraging model optimization techniques such as ONNX conversion, quantization, and TensorRT acceleration. The Jetson Orin Nano demonstrated the most balanced performance among all evaluated devices, confirming its suitability for deployment of the proposed real-time ADAS framework in edge environments.
 
\begin{table}[!h]
\centering
\caption{Performance of Lane Departure Warning, Traffic Sign Recognition, and Combined Task on different hardware platforms. The module marked with \textsuperscript{*} in the ``Framework'' column for the Jetson Orin Nano denotes that the model was run on CPU.}
\label{tab:combined_results}
\small
\begin{tabular}{l c c c c}
\toprule
\textbf{Hardware} & \textbf{Task} & \textbf{Framework} & \textbf{Processing Unit} & \textbf{Real-Time FPS} \\
\midrule
 
\multirow{6}{*}{Jetson Orin Nano}
 & \multirow{4}{*}{Lane} & PyTorch   & \multirow{6}{*}{GPU} & 25.50 \\
 &                       & ONNX      &  & 32.00 \\
 &                       & TensorRT  &  & 32.50 \\
 &                       & Quantized\textsuperscript{*} &  & 22.0 \\
 & Sign                  & ONNX      &  & 18.00 \\
 & Lane+Sign             & ONNX      &  & 15.00 \\
\midrule
 
Jetson Xavier   & Lane & PyTorch & GPU & 16.0 \\
\midrule
 
\multirow{3}{*}{Jetson Nano}
 & \multirow{3}{*}{Lane} & PyTorch   & \multirow{3}{*}{GPU} & 11.10 \\
 &      & ONNX      &  & 8.20  \\
 &      & TensorRT  &  & 11.40 \\
\midrule
 
\multirow{2}{*}{Raspberry Pi 4B}
 & \multirow{2}{*}{Lane} & PyTorch   & \multirow{2}{*}{CPU} & 0.22 \\
 &      & ONNX      &  & 1.80  \\
\midrule
 
\multirow{5}{*}{%
  \begin{tabular}[l]{@{}l@{}}
  Dell Laptop core i5 8\textsuperscript{th} Gen,\\
  16 GB RAM, 3.6 GHz processor \\
  without GPU
  \end{tabular}
}
 & \multirow{3}{*}{Lane} & PyTorch   & \multirow{5}{*}{CPU} & 0.37 \\
 &                       & ONNX      &  & 4.20  \\
 &                       & Quantized &  & 9.50  \\
 & Sign                  & ONNX      &  & 7.00  \\
 & Lane+Sign             & ONNX      &  & 5.50  \\
\bottomrule
\end{tabular}
\end{table}
 
\end{document}